\documentclass[10pt,twocolumn,letterpaper]{article}

\usepackage[pagenumbers]{cvpr}      
\usepackage{amsmath}
\usepackage{multirow}
\usepackage{graphicx}
\usepackage{geometry}
\usepackage{color}
\usepackage{xcolor,soul}
\usepackage{colortbl}
\definecolor{lightorange}{rgb}{0.98, 0.75, 0.5}
\definecolor{oorange}{RGB}{215,122,71}
\definecolor{yyellow}{RGB}{230,185,79}
\definecolor{ppurple}{RGB}{122,30,97}
\definecolor{ggreen}{RGB}{112,173,71}
\definecolor{lightblue}{rgb}{0.85, 0.95, 0.99}
\definecolor{battleshipgrey}{rgb}{0.3, 0.3, 0.3}
\definecolor{brilliantrose}{rgb}{1.0, 0.33, 0.64}
\definecolor{americanrose}{rgb}{1.0, 0.01, 0.24}
\definecolor{jweigreen}{rgb}{0,0.45,0.24}
\definecolor{jweired}{rgb}{0.45,0,0}
\definecolor{yellowish}{rgb}{0.8, 0.8, 0}
\definecolor{ao(english)}{rgb}{0.0, 0.5, 0.0}
\definecolor{blanchedalmond}{rgb}{1.0, 0.92, 0.8}
\definecolor{atomictangerine}{rgb}{1.0, 0.6, 0.4}
\definecolor{chocolate(web)}{rgb}{0.82, 0.41, 0.12}
\definecolor{bananayellow}{rgb}{1.0, 0.88, 0.21}
\definecolor{goldenbrown}{rgb}{0.6, 0.4, 0.08}
\definecolor{aliceblue}{rgb}{0.94, 0.97, 1.0}
\definecolor{beige}{rgb}{0.96, 0.96, 0.86}
\definecolor{babyblue}{rgb}{0.54, 0.81, 0.94}
\definecolor{camel}{rgb}{0.76, 0.6, 0.42}
\definecolor{cinnamon}{rgb}{0.82, 0.41, 0.12}
\definecolor{darkgreen}{rgb}{0.0, 0.4, 0.2}
\definecolor{darkred}{rgb}{0.55, 0.0, 0.0}
\definecolor{redlinkcolor}{rgb}{0.79607843, 0.25098039, 0.25882353}
\definecolor{bluecitecolor}{rgb}{0,0.36,0.69}
\definecolor{cvprblue}{rgb}{0.21,0.49,0.74} 
\usepackage{url}  
\usepackage{pifont} 
\newcommand{\huggingface}{\raisebox{-1.5pt}{\includegraphics[height=1.05em]{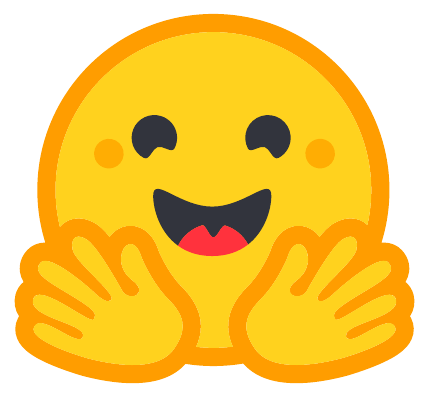}}\xspace}
\newcommand{\github}{\raisebox{-1.5pt}{\includegraphics[height=1.05em]{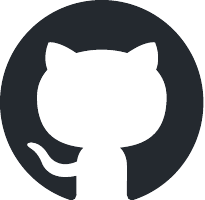}}\xspace}
\newcommand{\logo}{\raisebox{-1.5pt}{\includegraphics[height=2.0em]{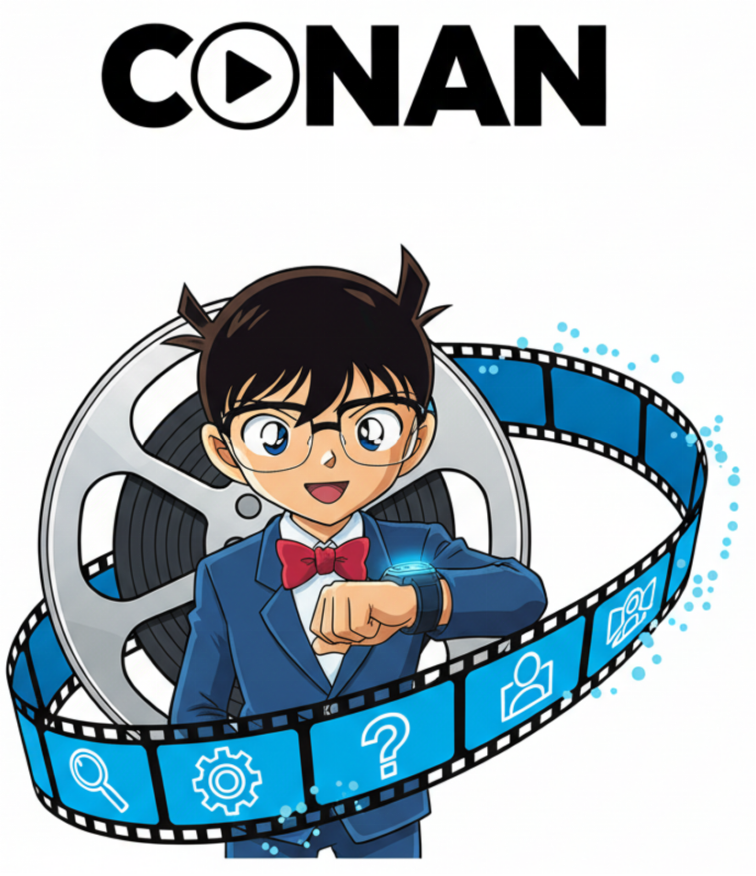}}\xspace}
\makeatletter
\newenvironment{chapquote}[2][2em]
  {\setlength{\@tempdima}{#1}%
   \def\chapquote@author{#2}%
   \parshape 1 \@tempdima \dimexpr\linewidth-2\@tempdima\relax%
   \itshape}
  {\par\normalfont\hfill--\ \chapquote@author\hspace*{\@tempdima}\par\bigskip}
\makeatother

\usepackage[pagebackref,breaklinks,colorlinks,allcolors=cvprblue]{hyperref}

\def\paperID{2479} 
\def\confName{CVPR}
\def\confYear{2026}

\title{\raisebox{-0.3\height}{\includegraphics[scale=0.1]{figure/logo.png}}\hspace{0.2em}Conan: Progressive Learning to Reason Like a Detective over Multi-Scale Visual Evidence}
\author{Kun Ouyang$^{1}$,\quad Yuanxin Liu$^{1}$,\quad Linli Yao$^{1}$,\quad Yishuo Cai$^{1}$,\\ Hao Zhou$^{2}$,\quad Jie Zhou$^{2}$,\quad Fandong Meng$^{2}$,\quad Xu Sun$^{1}$\thanks{Corresponding Author}\\
$^{1}$State Key Laboratory for Multimedia Information Processing, 
\\School of Computer Science, Peking University\\
$^{2}$WeChat AI, Tencent Inc., China\\
{\tt\small kunouyang10@gmail.com,}\quad {\tt\small xusun@pku.edu.cn}
}

\begin{document}
\makeatletter
\let\@oldmaketitle\@maketitle%
\renewcommand{\@maketitle}{\@oldmaketitle%
\vspace{-2em}
    \centering
    \includegraphics[width=.78\textwidth]{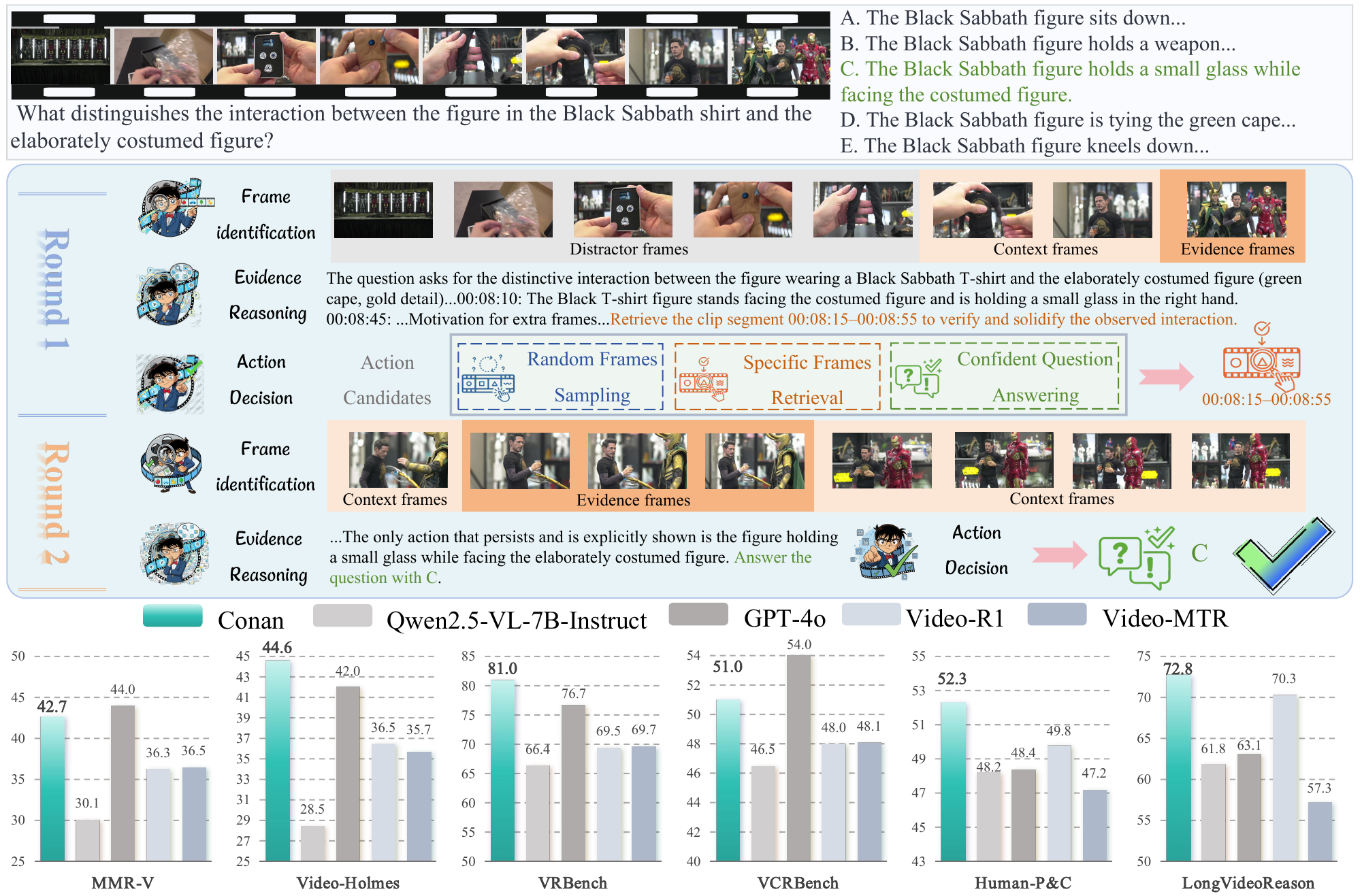}
    \captionof{figure}{\textbf{Top}: The reasoning process of our Conan on an example question. \textbf{Bottom}: Performance comparison across six multi-step reasoning benchmarks.
    }
    \label{fig:teaser}
   \bigskip}%
\makeatother
\maketitle

\begin{abstract}
Video reasoning, which requires multi-step deduction across frames, remains a major challenge for multimodal large language models (MLLMs). While reinforcement learning (RL)-based methods enhance reasoning capabilities, they often rely on text-only chains that yield ungrounded or hallucinated conclusions. Conversely, frame-retrieval approaches introduce visual grounding, yet still struggle with inaccurate evidence localization. To address these limitations, we present Conan, a framework for evidence-grounded multi-step video reasoning. Conan identifies context and evidence frames, reasons over cross-frame clues, and adaptively decides when to conclude or explore further. To achieve this, we \textbf{1)} construct Conan-91K, a large-scale dataset of automatically generated reasoning traces that include frame identification, evidence reasoning, and action decision, and \textbf{2)} design a multi-stage progressive cold-start strategy combined with an Identification-Reasoning-Action (AIR) RLVR training framework to progressively incentivize multi-step visual reasoning. Extensive experiments on six multi-step reasoning benchmarks demonstrate that Conan surpasses the baseline Qwen2.5-VL-7B-Instruct by an average of over 10\% in accuracy, achieving state-of-the-art performance. Furthermore, Conan generalizes effectively to long video understanding tasks, validating its strong scalability and robustness.

\vspace{-10pt}
\renewcommand{\arraystretch}{1.2}
\begin{center}
\resizebox{\linewidth}{!}{
\begin{tabular}{cll}
\logo & \textbf{Model:} &  \href{https://huggingface.co/RUBBISHLIKE/Conan-7B} {\path{huggingface.co/RUBBISHLIKE/Conan-7B}}\\
\huggingface & \textbf{Dataset:} &  \href{https://huggingface.co/RUBBISHLIKE/Conan-91k} {\path{huggingface.co/RUBBISHLIKE/Conan-91k}}\\
\github & \textbf{Code:} & \href{https://github.com/OuyangKun10/Conan}{\path{github.com/OuyangKun10/Conan}}\\
\\
\end{tabular}
}
\end{center} 
\vspace{-12pt}
\end{abstract}    
\begin{figure*}[htb]
    \centering
    \includegraphics[width=0.9\textwidth]{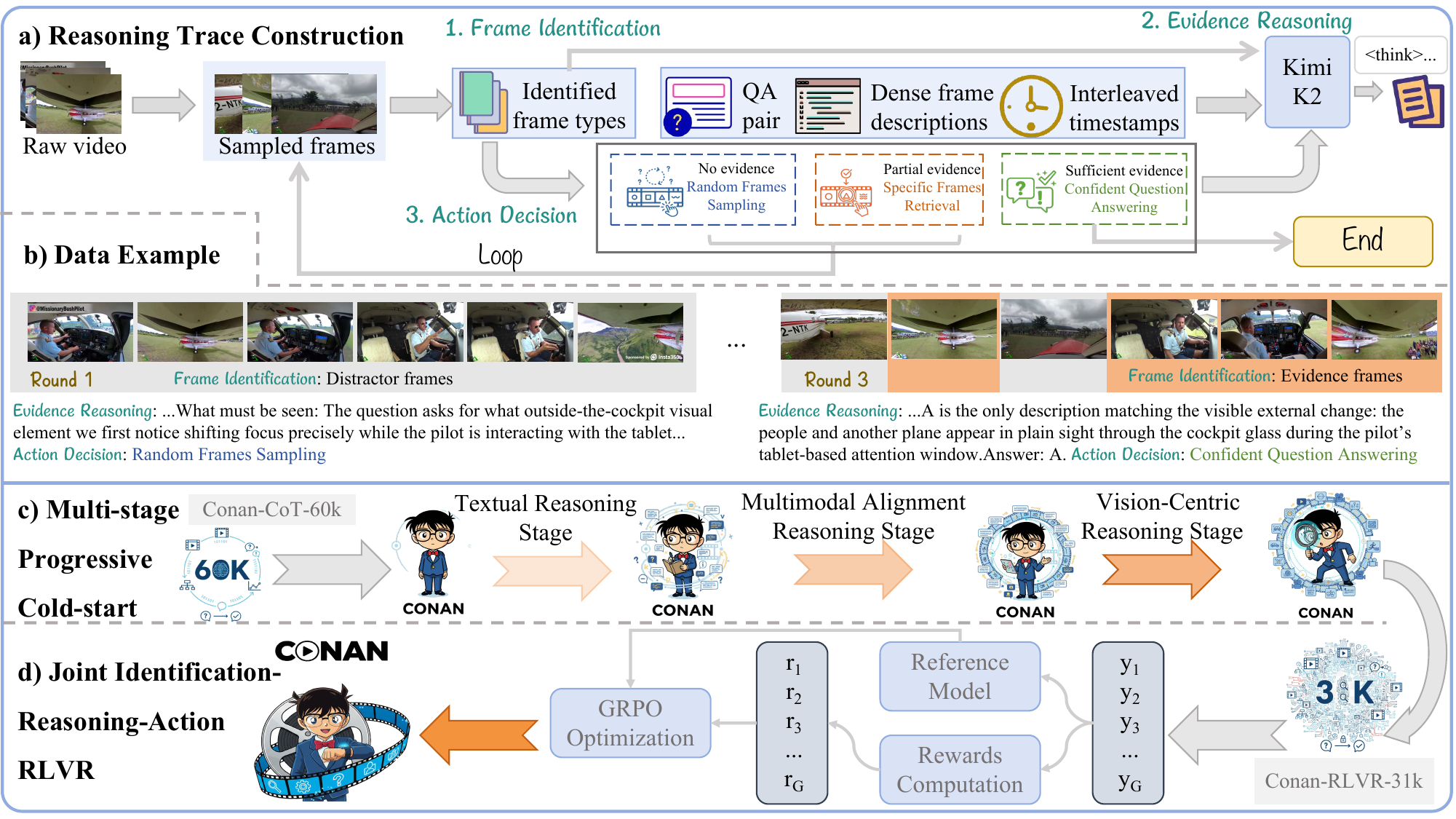}
     \caption{\textbf{a)} Reasoning Trace Construction. \textbf{b)} Data Example.
     \textbf{c)} Multi-stage Progressive Cold-start, including textual, multimodal alignment, and vision-centric reasoning stages. \textbf{d)} The Joint Identification-Reasoning-Action RLVR.}
    \label{Fig:data_example}
\end{figure*}
\section{Introduction}
\label{sec:intro}
\begin{chapquote}{Edogawa Conan}
There is only one truth!
\end{chapquote}
Frontier multimodal large language models (MLLMs)~\cite{Qwen2.5-VL,kimiteam2025kimivltechnicalreport,coreteam2025mimovltechnicalreport,GPT-4o} have demonstrated remarkable progress on standard video understanding tasks such as question answering~\cite{vqaiccv}, temporal grounding~\cite{gao2017talltemporalactivitylocalization}, and captioning~\cite{venugopalan2015translating}. However, video reasoning~\cite{Cheng2025VideoHolmesCM,Liu2025VideoReasonBenchCM} remains a substantial challenge. Unlike conventional tasks, video reasoning~\cite{Zhu2025MMRVWL,Cheng2025VideoHolmesCM,Yu2025VRBenchAB,qi2025vcr,chen2025longvila-r1} demands active visual information accumulation across temporal spans and multi-step logical inference to reach well-grounded conclusions.

Inspired by the success of reinforcement learning with verifiable rewards~\cite{deepseekai2025deepseekr1incentivizingreasoningcapability} (RLVR) in incentivizing reasoning ability of LLMs, recent works~\cite{feng2025videor1reinforcingvideoreasoning,Ouyang2025SpaceRRM,li2025videochatr1enhancingspatiotemporalperception} have begun extending this paradigm to video reasoning, achieving promising gains. Nevertheless, these approaches primarily rely on pure-text reasoning without explicit grounding in visual evidence, often leading to superficial or hallucinated reasoning chains that fail to reflect the actual video content.
To integrate visual evidence into the reasoning process, concurrent works~\cite{Xie2025VideoMTRRM,zhang2025rewatch,he2025framethinker} introduce frame retrieval mechanism to enable video chain-of-thought (Video-CoT) reasoning, boosting performance of long video understanding~\cite{wu2024longvideobench,Zhou2024MLVUBM}. However, these approaches usually suffer from inaccurate or implicit evidence localization, yielding unreliable reasoning paths. Additionally, some methods~\cite{he2025framethinker,zhang2025rewatch} partially rely on benchmark-specific training data (\eg Video-Holmes~\cite{Cheng2025VideoHolmesCM} and LongVideoReason~\cite{chen2025longvila-r1}), making it difficult to disentangle solid reasoning improvements from in-domain overfitting.

Motivated by these limitations, we aim to equip MLLMs with multi-step, evidence-grounded video reasoning skills, analogous to how Conan acts as a detective (Figure~\ref{fig:teaser}): Specifically, our framework identifies relevant frames at multiple scales (context and evidence frames), reasons over cross-frame clues to form coherent chains of deduction, and decides whether to draw the final conclusion or continue exploring the video. Achieving this goal raises two core challenges: 1) \textbf{How to automatically construct a high-quality evidence-based reasoning dataset} that explicitly captures evidence localization, multi-step deductive reasoning, and confident action decision. 2) \textbf{How to design training curriculum for effective acquisition of visual reasoning across multi-scale evidence}.

To tackle the \textbf{first} challenge, we introduce Conan-91k, a large-scale dataset for Conan-style evidence reasoning. Built upon the key-frame identification dataset GenS-Video-150K~\cite{yao-etal-2025-generative}, we develop an automated pipeline to generate interleaved video-text reasoning traces using the advanced LLM Kimi K2~\cite{team2025kimi}, as illustrated in Figure~\ref{Fig:data_example}. Each reasoning trace contains three key components: a) \textbf{Frame Identification} distinguishes among evidence, context, and distractor frames. b) \textbf{Evidence Reasoning} conducts textual reasoning over the question and accumulated visual clues. c) \textbf{Action Decision} decides whether to retrieve additional frames or reach the final conclusion. Furthermore, we propose an evidence difficulty-aware sampling strategy to facilitate progressive training from simple to complex reasoning, producing 60k samples (Conan-CoT-60k) for SFT and 31k samples (Conan-RLVR-31k) for RL.

To address the \textbf{second} challenge, we introduce a two-phase training  curriculum: a) A multi-stage progressive cold-start strategy incrementally activates the model's multi-step reasoning using Conan-CoT-60k, starting from textual reasoning, advancing to multimodal alignment, and culminating in vision-centric deduction. b) On this foundation, the joint Identification-Reasoning-Action (AIR) RLVR framework further refines the model's capacity to perform Conan-style reasoning on Conan-RLVR-31k by jointly optimizing frame identification, evidence reasoning, and action decision.
Together, these components empower Conan to ``seek, deduce, and act'' across visual clues, achieving reliable multi-step reasoning.

Extensive experiments on six challenging multi-step reasoning benchmarks (\eg MMR-V~\cite{Zhu2025MMRVWL}, Video-Holmes~\cite{Cheng2025VideoHolmesCM}, VRBench~\cite{Yu2025VRBenchAB}, VCRBench~\cite{qi2025vcr}, Human-P\&C~\cite{li2025humanpcrprobingmllmcapabilities}, and LongVideoReason~\cite{chen2025longvila-r1}) demonstrate that Conan consistently surpasses the base model Qwen2.5-VL-7B-Instruct~\cite{Qwen2.5-VL}, with an average accuracy improvement exceeding 10\%. Moreover, Conan achieves promising enhancement on long video understanding tasks (\eg LongVideoBench~\cite{wu2024longvideobench}, MLVU~\cite{Zhou2024MLVUBM}, LVBench~\cite{wang2024lvbench}, and Video-MME~\cite{fu2024video}), validating strong scalability and robustness.

In summary, our contributions are threefold:
\ding{171} We introduce Conan-91k, the first large-scale dataset for multi-scale evidence reasoning, including Conan-CoT-60k and Conan-RLVR-31k.
\ding{170} We propose a multi-stage progressive cold-start strategy and a joint AIR RLVR framework to foster gradual acquisition of evidence-based multi-step reasoning skills.
\ding{168} We conduct extensive experiments, demonstrating that Conan achieves substantial improvements in both multi-step reasoning and long video understanding.

\section{Dataset Construction}
\vspace{-0.1cm}
\subsection{Data Collection \& Processing}
We collect source data from the GenS-Video-150K~\cite{yao-etal-2025-generative} dataset, which provides dense frame descriptions, multi-choice and free-form QA pairs, as well as frame-level relevance scores. Leveraging these relevance scores, we categorize video frames into three types: 1) Evidence frames, which are directly relevant to answering the question;
2) Context frames, which offer auxiliary hints that may support the reasoning process; and
3) Distractor frames, which bear no relation to the question.
This multi-scale frame categorization establishes the foundation for subsequent stepwise reasoning trace construction.
\subsection{Reasoning Trace Construction}
Starting from the processed data, we apply an automatic pipeline to construct Conan-style video-text interleaved reasoning traces with the assistance of a strong LLM, \emph{Kimi K2}~\cite{team2025kimi}, as illustrated in Figure~\ref{Fig:data_example}. The core loop proceeds as follows:
\begin{itemize}
    \item We first sample $16$ frames uniformly from the raw video and retain each frame's type label (evidence, context, or distractor).
    \item If all frames are distractors, we select the action \textit{Random Frames Sampling}, which randomly samples $8$ new frames and continue the loop.
    \item If some sampled frames are evidence or context but the evidence proportion does not exceed a dynamic answering threshold at which K2 can reach the answer based on the acquired clues, we select the action \textit{Specific Frames Retrieval}, which uniformly retrieves $8$ frames within a single clip or multiple clips that contain evidence/context, and continue the loop.
    \item If the proportion of evidence frames exceeds the threshold, we terminate the loop by selecting \textit{Confident Question Answering} as the final action.
\end{itemize}
For every loop iteration, we prompt K2 with the chosen action, the frame types, the QA pair, dense frame descriptions, and timestamps to generate a coherent textual reasoning trace that a) analyzes the QA and sampled frames, and b) justifies the chosen action. More details are in Appendix~B.

\subsection{Evidence Difficulty-Aware Sampling}
To facilitate a progressive training curriculum from simple to complex reasoning cases, we propose an evidence difficulty-aware sampling 
strategy. In particular, we define an Evidence Difficulty Index (EDI) to quantify the reasoning complexity based on the proportion and temporal dispersion of evidence frames.
Let the evidence ratio be $P=m/N$, where $m$ and $N$ denote the numbers of evidence and total frames, respectively. The temporal variance of evidence $\mathrm{Var}= \frac{1}{m} \sum_{i=1}^{m} \left(x_i - \bar{x}\right)^2$,
where $x_i$ represents the temporal position of $i$-th evidence frame, and $\bar{x}$ is the mean position of all evidence frames.
The overall difficulty is defined as $\text{EDI}= (1-P)\,\mathrm{Var}$, where higher EDI value indicates sparser and more temporally dispersed evidence, reflecting greater difficulty.

Based on the EDI distribution and reasoning round, samples are stratified and adaptively allocated across training phases. During the \textbf{SFT} stage, we focus on lower-EDI samples to emphasize foundational reasoning skills, selecting 60k instances with up to three reasoning rounds: 25k one-round, 25k two-round, and 10k three-round samples that align with the overall round distribution. In contrast, the \textbf{RLVR} stage employs higher-EDI samples without constraining reasoning rounds, selecting 31k instances spanning diverse reasoning rounds.
This difficulty-aware sampling scheme establishes a principled curriculum that transitions smoothly from low-difficulty grounding in SFT to high-difficulty multi-hop reasoning in RLVR, facilitating the gradual and robust acquisition of evidence-based reasoning capabilities. The resulting datasets, \textbf{Conan-CoT-60k} and \textbf{Conan-RLVR-31k}, together form the \textbf{Conan-91k} dataset. Dataset statistics and additional construction details are provided in Appendix~B.
\section{Two-Phase Training Curriculum}
\subsection{Multi-Stage Progressive Cold-Start}
To progressively activate the multi-step reasoning abilities, we conduct a multi-stage progressive cold-start on Conan-CoT-60k, guided by a stage-wise incremental sampling that gradually expands data diversity and reasoning difficulty:
\textbf{1) Textual Reasoning Stage.} In the initial stage, the model is trained on $10$k relatively low-EDI samples from one-round subset, where frames are represented by dense textual descriptions and timestamps. This stage focuses on temporal and causal reasoning across ordered frame descriptions, establishing a structured reasoning foundation for subsequent multimodal learning.
\textbf{2) Multimodal Alignment Reasoning Stage.} Compared with the first stage, we utilize $25$k one-round samples (including $15$k new ones) and a new set of $10$k relatively low-EDI samples from two-round subset, inserting timestamps and textual descriptions before visual frames. This incremental expansion ensures that the model continues learning on partially new data, preventing overfitting to previously seen questions while promoting stable adaptation from textual to multimodal reasoning. The two-round samples allow the model to collect more evidence for reasoning.
\textbf{3) Vision-Centric Reasoning Stage.} In the final stage, the model is trained on the complete Conan-CoT-60k, including additional $15$k two-round and $10$k three-round samples. This stage compels the model to execute deep multi-step reasoning directly over visual frames with interleaved timestamps, thereby enhancing perceptual grounding and fostering vision-centric reasoning.

\subsection{AIR RLVR}
Building upon the model Conan-SFT obtained from the previous cold-start process, we further refine its multi-step reasoning capabilities via the Identification-Reasoning-Action (AIR) RLVR framework.
Given that the model has already learned to produce reasoning traces consisting of: 1) frame identification, 2) evidence reasoning, and 3) action decision, AIR RLVR aims to optimize the exploration of effective reasoning trajectories through a set of carefully designed reward functions.
We first introduce one format reward and two outcome rewards to ensure both structural consistency and answer accuracy.

\noindent\textbf{Format Reward.} To enforce structural consistency in model outputs $y$, we define a format reward $R_{fmt}$ that verifies whether specific tags are correctly applied. And the model is restricted to performing only one action (\eg random frames sampling, specific frames retrieval, confident question answering) at a time. The format reward $R_{fmt}$ is defined as:
\begin{equation} \label{eq:fmt_reward}
R_{fmt}(y) = 
\begin{cases} 
0.5, & \text{if $y$ matches format}, \\
0, & \text{otherwise}.
\end{cases}
\end{equation}

\noindent\textbf{Multi-choice Reward.} For multi-choice QA, the outcome reward $R_{mc}$ is determined by exact match between the predicted answer $y$ and ground truth $\hat{y}$:
\begin{equation} \label{eq:multi-choice_reward}
R_{mc}(y,\hat{y}) = 
\begin{cases} 
1, & \text{if } y = \hat{y}, \\
0, & \text{otherwise}.
\end{cases}
\end{equation}

\noindent\textbf{Free-form Reward.} For free-form QA, the outcome reward $R_{free}$ is computed as the average of ROUGE-1, ROUGE-2, and ROUGE-L scores~\cite{Lin2004ROUGEAP} between the predicted answer $y$ and the ground truth $\hat{y}$:
\begin{equation}\label{eq:free_reward}
R_{free}(y,\hat{y})
= \frac{1}{3}\bigl(
\operatorname{R_1}(y,\hat{y})
+\operatorname{R_2}(y,\hat{y})
+\operatorname{R_L}(y,\hat{y})
\bigr)
\end{equation}

To evaluate the quality of the multi-scale frames identification, and frames retrieval actions, we design an identification reward $R_{ide}$ and a retrieval reward $R_{ret}$, respectively: 1) The \textbf{identification reward} $R_{ide}$ measures the average accuracy of identified evidence/context frames across reasoning rounds.
2) The \textbf{retrieval reward} $R_{ret}$ evaluates the quality of retrieved frames by computing the average ratio of evidence/context frames among all retrieved frames.
And the final joint identification-retrieval-outcome reward $R_{J}$ is formulated as:
\begin{equation}\label{eq:ERO_reward}
R_{J} =\left\{
\begin{aligned}
&R_{fmt} + R_{o} + R_{ide} + R_{ret}, &\text{if } R_{o} > 0, \\
&R_{fmt} + R_{o}, &\text{otherwise}, 
\end{aligned}
 \right.
\end{equation}
where $R_{o}$ is the outcome reward and $o \in \{mc, free\}$.
This reward shaping encourages the model to generate structurally valid, evidence-grounded, and accurate reasoning traces while improving retrieval efficiency. Finally, we prompt the model generate a group of responses $\{y_1, y_2, \cdots, y_G\}$, where $G$ is the number of generated responses, and we adopt the GRPO~\cite{deepseekai2025deepseekr1incentivizingreasoningcapability} algorithm for reinforcement optimization to stabilize training and refine the reasoning policy.

\begin{table*}[ht]
    \centering
    \resizebox{\textwidth}{!}{
        \begin{tabular}{l|c|c|c|c|c|c|c|c|c}
        \toprule
            & \#Params & \textbf{MMR-V} & \textbf{Video-Holmes} & \textbf{VRBench} & \textbf{VCRBench$\ast$}&  \textbf{Human-P\&C}&\textbf{LVR}   & \textbf{Avg. VR} & \textbf{Avg. LU} \\
        \midrule
            \rowcolor{lightgray}\multicolumn{10}{l}{\textit{Closed-source Models}} \\
            GPT-4o~\cite{GPT-4o} &-&44.0&42.0&76.7&54.0&48.4&63.1&54.7&-\\
            Gemini 1.5 Pro&-&-&41.2&-&-&52.6&69.3&-&-\\
            Gemini 2.0 Flash &-&42.6&30.6&-&-&56.1&-&-&-\\
              \midrule
          \rowcolor{lightgray}\multicolumn{10}{l}{\textit{Open-source Models}} \\
          LLaVA-OneVision-7B~\cite{li2024llava}
          &7B&6.5&-&-&30.7&48.5&-&-&-\\
         
        InternVL3-8B~\cite{chen2024expanding}
        &8B&35.5&32.8&75.8&45.7&51.0&68.0&51.5&52.5\\
         Kimi-VL-A3B-Instruct~\cite{kimiteam2025kimivltechnicalreport} 
         &3B/16B  &32.4 &32.4&60.5&34.3&42.0&64.6&44.4&- \\
           Qwen2.5-VL-72B-Instruct~\cite{Qwen2.5-VL}
           &72B &39.1&40.2&72.7&50.8&55.7&72.3&55.1&53.4\\
          \rowcolor{aliceblue}Qwen2.5-VL-7B-Instruct~\cite{Qwen2.5-VL} 
          & 7B  &30.1&28.5&66.4&46.5&48.2&61.8& 46.9&48.0 \\ 
        \rowcolor{lightgray}\multicolumn{10}{l}{\textit{Text CoT Models}} \\
        Video-R1~\cite{feng2025videor1reinforcingvideoreasoning}
        &7B&36.3&36.5&69.5&48.0&49.8&70.3&51.7&53.8 \\
        VideoChat-R1~\cite{li2025videochatr1enhancingspatiotemporalperception}
        &7B&36.1&33.0&61.5&48.2&51.8&67.9&49.8&52.4\\
        \rowcolor{lightgray}\multicolumn{10}{l}{\textit{Video CoT Models}} \\
        Video-MTR~\cite{Xie2025VideoMTRRM} \textcolor{gray}{Concurrent work}
        &7B&36.5&35.7&69.7&48.1&47.2&57.3&49.1&53.5\\
        Rewatch-R1$\dag$~\cite{zhang2025rewatch} \textcolor{gray}{Concurrent work}
        &7B&45.3&37.8&79.1&49.8&51.6&70.5&55.7&50.5\\
   
          \midrule
         Conan \textcolor{gray}{SFT}&7B&35.4&34.9&64.4&43.3&50.4&66.0&49.1&49.3\\ 
            \rowcolor{lightblue} Conan& 7B    &{42.7}~\textcolor{jweigreen}{ \small($\uparrow 12.6$ )}  & {44.6}~\textcolor{jweigreen}{ \small($\uparrow 16.1$ )}  & {81.0}~\textcolor{jweigreen}{ \small($\uparrow 14.6$ )} & {51.0}~\textcolor{jweigreen}{ \small($\uparrow 4.5$ )}&{52.3}~\textcolor{jweigreen}{ \small($\uparrow 4.1$ )} & {72.8}~\textcolor{jweigreen}{ \small($\uparrow 11.0$ )} & {57.4} ~\textcolor{jweigreen}{ \small($\uparrow 10.5$ )} &{54.9}~\textcolor{jweigreen}{ \small($\uparrow 6.9$ )}\\ 
        \bottomrule
        \end{tabular}
    }
    \caption{Main results of \colorbox{aliceblue}{Qwen2.5-VL-7B-Instruct}, \colorbox{lightblue}{Conan}, and other baselines. \textbf{Avg. VR} and \textbf{Avg. LU} denote the average results on six video reasoning benchmarks and four long video understanding benchmarks, respectively.
    $\dag$ indicates partial training on the training set of LongVideoReason (LVR)~\cite{chen2025longvila-r1}, and $\ast$ marks the multiple-choice subset.}
    \label{Tab:main_exp}
\end{table*}
\section{Experiment}
\subsection{Evaluation Setups}

\noindent\textbf{Implementation Details}
1) Training. We adopt Qwen-2.5-VL-7B-Instruct~\cite{Qwen2.5-VL} as the base model.
During the cold start, the model is trained for up to one epoch per stage with a global batch size of $32$ and a learning rate of $1e$-$5$. The trained model is then used for the AIR RLVR phase, also trained for one epoch under the batch size of $24$ and the learning rate of $1e$-$6$. The maximum completion length is set to $4,000$ tokens, with a generation temperature of $1.0$ and a generation number $G$ of $8$.
Each input video contains $16$ initial frames, and the model can retrieve up to $8$ additional frames per reasoning step.
2) Evaluation. The generation temperature is fixed at $1.0$, and each sample is evaluated three times. The maximum number of new tokens is set to $4,000$ when reasoning traces are included and $128$ for direct answering. The reasoning process is limited to three rounds. Videos are standardized to $16$ frames for multi-step reasoning and $32$ frames for long video understanding during evaluation, with a resolution of $448\times28\times28$.
\\
\noindent\textbf{Benchmarks \& Baselines.}
We conduct comprehensive evaluations on six multi-step reasoning benchmarks(\ie, MMR-V~\cite{Zhu2025MMRVWL}, Video-Holmes~\cite{Cheng2025VideoHolmesCM}, VRBench~\cite{Yu2025VRBenchAB}, VCRBench~\cite{qi2025vcr} (multi-choice subset), Human-P\&C~\cite{li2025humanpcrprobingmllmcapabilities}, and LongVideoReason~\cite{chen2025longvila-r1}) and four long video understanding benchmarks (\ie, LongVideoBench~\cite{wu2024longvideobench}, MLVU~\cite{Zhou2024MLVUBM}, LVBench~\cite{wang2024lvbench}, and Video-MME~\cite{fu2024video}), comparing Conan with a range of closed-source and open-source MLLMs. Descriptions of benchmarks and baselines are provided in Appendix~C.1.

\subsection{Main Results}
The main evaluation results are shown in Table~\ref{Tab:main_exp}, while the full results of long video understanding are in Table~$3$ of Appendix.
The key observations are summarized as follows.
\begin{figure}[htb]
    \centering
    \includegraphics[width=0.9\linewidth]{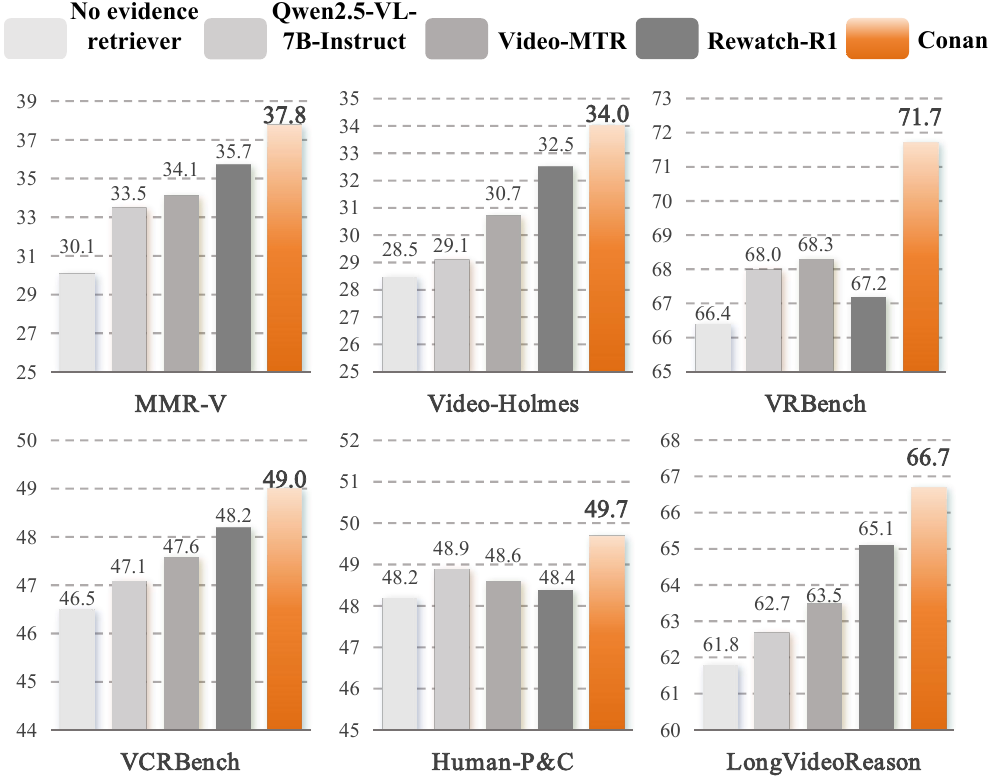}
     \caption{Performance comparison across different evidence retrievers, with Qwen2.5-VL-7B-Instruct as the question-answering model.}
    \label{Fig:sampler}
    \vspace{-0.6cm}
\end{figure}
\begin{table*}[ht]
    \centering
    \resizebox{\textwidth}{!}{
        \begin{tabular}{l|c|c|c|c|c|c|c}
        \toprule
           & \textbf{MMR-V} & \textbf{Video-Holmes} & \textbf{VRBench} & \textbf{VCRBench$\star$} &  \textbf{Human-P\&C}&\textbf{LVR}&\textbf{Overall}\\
          \midrule
     \rowcolor{lightgray}\multicolumn{8}{l}{\textit{Dataset design}} \\
     w/o multi-scale evidence &39.7&39.1&75.4&46.9&51.3&70.2&53.8\\
     w/o difficulty sampling  &40.0&40.8&78.9&48.4&50.7&72.3&55.2\\
     \midrule
     \rowcolor{lightgray}\multicolumn{8}{l}{\textit{Progressive cold-start}} \\
     w/o textual reasoning  &42.2  & 43.9 & \textbf{81.4} & 49.8&51.7 &\textbf{73.1}&57.0 \\ 
     w/o multimodal alignment reasoning  &\textbf{43.5}  &44.2  & 80.4 & 48.2 &49.3 & 72.5  &56.4\\ 
     w/o vision-centric reasoning  &39.0  & 36.5 & 75.2 & 50.2 &48.7 & 68.5 &53.0\\
     w/o cold-start  &39.1  &36.8  &73.3  &46.3  &47.9&62.3 &51.0\\ 
     \midrule
     \rowcolor{lightgray}\multicolumn{8}{l}{\textit{AIR RLVR}} \\
     w/o identification reward  &39.9  & 40.0 & 74.8 &46.1   &50.1&71.7&53.8\\ 
     w/o retrieval reward  &38.2  & 42.4 & 75.7 & 47.8 &50.5& 69.2&54.0\\ 
     w/ text CoT &41.1&41.0&76.8&48.8&\textbf{52.8}&70.5&55.2\\
     \midrule
     Conan &42.7  & \textbf{44.6} &81.0 & \textbf{51.0} &52.3& 72.8 &\textbf{57.4}\\ 
        \bottomrule
        \end{tabular}
    }
    \caption{Ablation results of Conan, where the best results are in \textbf{boldface}.}
    \label{Tab:ablation_exp}
    \vspace{-0.3cm}
\end{table*}
\\ \noindent \textbf{Overall Analysis.}
Across video reasoning benchmarks, Conan substantially surpasses its base model Qwen2.5-VL-7B-Instruct with an average accuracy gain of over $10$\%. Remarkably, Conan also outperforms the advanced GPT-4o on most benchmarks (\eg, Video-Holmes, VRBench, Human-P\&C, and LongVideoReason), underscoring its superior capabilities of multi-step, evidence-grounded reasoning. And the two text-CoT models (\eg, Video-R1 and VideoChat-R1) perform notably worse than Conan, demonstrating the advantages of visual evidence identification and retrieval. Moreover, Conan consistently outperforms video-CoT approaches (\eg, Video-MTR and Rewatch-R1), showcasing more precise evidence localization, and more effective frame retrieval of Conan. Notably, Conan also generalizes well to long video understanding tasks, with an average improvement of $6.9$\%, proving the robustness and scalability of our framework.
\\
\noindent\textbf{Conan as an Evidence Retriever.}
To further assess Conan's evidence localization capability, we compare different models used as evidence retriever, each selecting up to 16 key frames per video for Qwen2.5-VL-7B-Instruct to perform question answering. As shown in Figure~\ref{Fig:sampler}, Conan yields the highest performance, highlighting its superior ability to identify informative visual evidence. Additional results under varying frame settings and base models are provided in Appendix~C.2. And training dynamics can be found in Appendix~C.3.

\begin{figure*}[htb]
    \centering
    \includegraphics[width=0.9\linewidth]{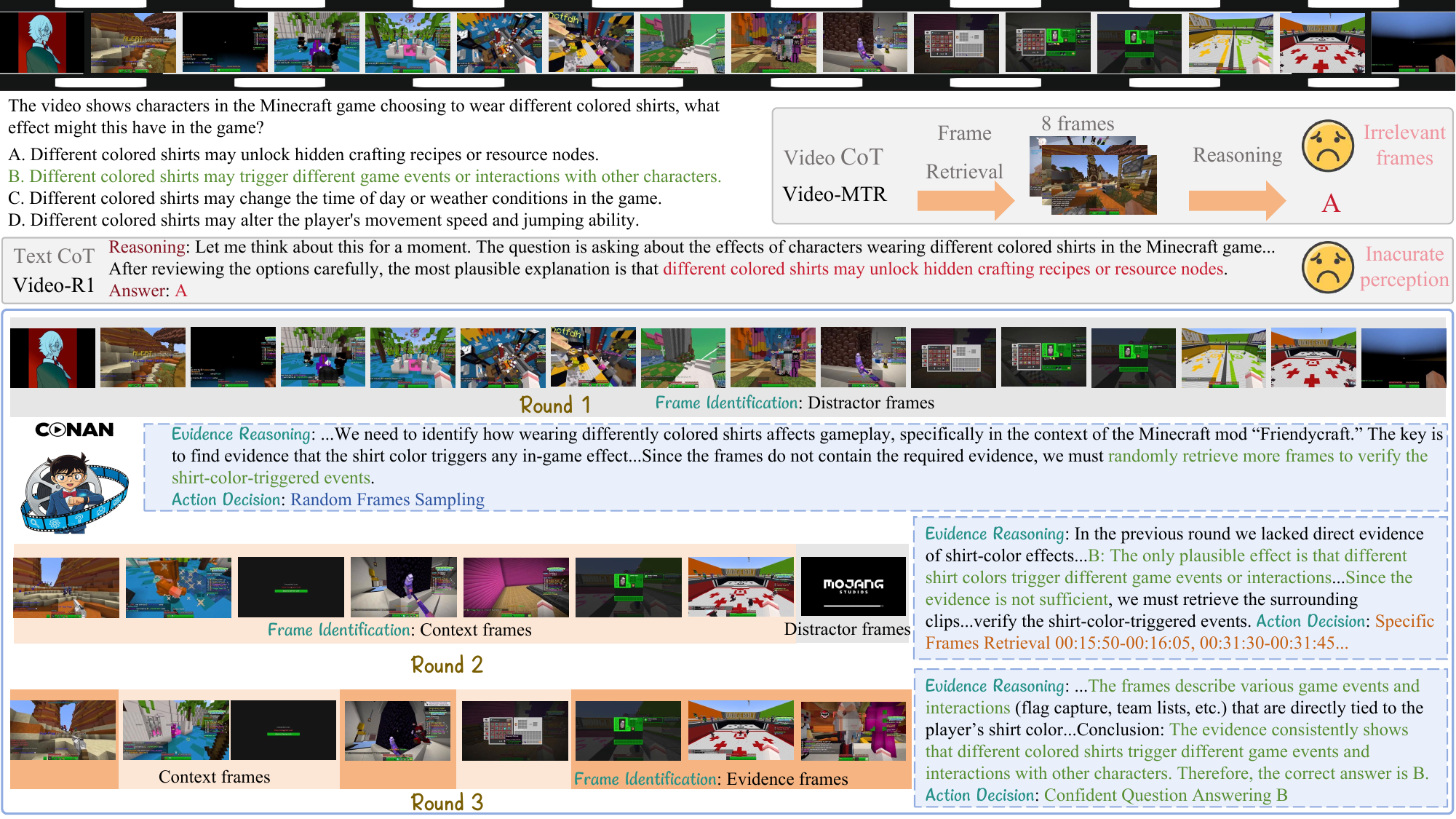}
     \caption{A qualitative example from VRBench, including the QA pair and the reasoning traces of Video-R1 (Text CoT), Video-MTR (Video CoT), and Conan.}
    \label{Fig:case1}
    \vspace{-0.3cm}
\end{figure*}
\subsection{Ablation Study}
To investigate the contribution of each component in our framework, we conduct a comprehensive ablation study with multiple Conan variants.
\\
\indent For the dataset design, we introduce two variants:
1) \textit{w/o multi-scale evidence},
which merges the \textit{context} type into the \textit{distractor} category in multi-scale frame categorization;
and
2) \textit{w/o difficulty sampling}, which replaces the evidence difficulty-aware sampling with random sampling.
\\
\indent For the multi-stage progressive cold-start strategy, we develop four variants:
1) \textit{w/o textual reasoning}, which excludes the textual reasoning stage;
2) \textit{w/o multimodal alignment reasoning}, which skips the multimodal alignment reasoning stage; and
3) \textit{w/o vision-centric reasoning}, which removes the vision-centric reasoning stage.
4) \textit{w/o cold-start}, which bypasses the three cold-start stages and directly employs AIR RLVR to train the model on the Conan-RLVR-31k dataset.
\\
\indent For the AIR RLVR framework, we design three variants:
1) \textit{w/o identification reward}, which discards the identification reward;
2) \textit{w/o retrieval reward}, which removes the retrieval reward; and
3) \textit{w/ text CoT}, which enforces single-round text-CoT paradigm performing pure textual reasoning and answering without additional retrieval in training and evaluation procedures.
\\
\indent The ablation results in Table~\ref{Tab:ablation_exp} reveal several key findings:
a) \textbf{Dataset design.} \textit{w/o multi-scale evidence} underperforms Conan, confirming the benefits of multi-scale frame identification in providing richer contextual cues. Moreover, Conan outperforms \textit{w/o difficulty sampling}, which validates that evidence difficulty-aware sampling effectively guides the model to progressively acquire multi-step reasoning abilities.
b) \textbf{Progressive cold-start.} Conan consistently surpasses \textit{w/o textual reasoning}, \textit{w/o multimodal alignment reasoning}, \textit{w/o vision-centric reasoning}, and \textit{w/o cold start}, which demonstrates that the multi-stage progressive cold-start is crucial for gradually activating the model's multi-hop reasoning capabilities. 
c) \textbf{AIR RLVR.} Conan outperforms \textit{w/o identification reward} and \textit{w/o retrieval reward}, proving their effectiveness in refining evidence localization accuracy and frame retrieval efficiency, respectively. And \textit{w/ text CoT} underperforms Conan, which validates the superiority of the Conan-style reasoning paradigm.

\subsection{Qualitative Evaluation}
Figure~\ref{Fig:case1} presents a qualitative comparison on VRBench~\cite{Yu2025VRBenchAB} among Video-R1, Video-MTR, and Conan.
Video-R1 conducts pure textual reasoning without visual grounding, resulting in a hallucinated answer driven by linguistic priors. Video-MTR incorporates frame retrieval but fails to localize relevant evidence, leading to weak alignment between retrieved frames and the question.
In contrast, Conan performs multi-step reasoning with accurate evidence grounding and efficient frame retrieval. In Round 1, it recognizes the absence of causal cues and performs random frames sampling to expand the search space. In Round 2, guided by contextual signals, it retrieves frames around key timestamps of player interactions. In Round 3, Conan identifies visual evidence of color-triggered game events (\eg, team activities, flag captures) and integrates these clues to infer the correct option.
This comparison highlights Conan's superior ability to localize, reason over, and act upon multi-scale visual evidence. More qualitative cases and error analyses are in Appendix~C.4.

\section{Conclusion and Future work}
In this work, we present Conan, a novel framework that equips MLLMs with Conan-like visual reasoning through frame identification, evidence reasoning, and action decision. Employing the Conan-91k dataset, constructed via multi-scale frame categorization, reasoning trace construction, and evidence difficulty-aware sampling, we devise a multi-stage progressive cold-start strategy alongside a joint Identification-Reasoning-Action (AIR) RLVR framework to progressively cultivate robust multi-step reasoning abilities. Extensive experiments across six multi-step reasoning and four long video understanding benchmarks demonstrate that Conan consistently outperforms the base model Qwen2.5-VL-7B-Instruct, achieving state-of-the-art performance over both text-CoT and video-CoT models.
In future work, we plan to extend Conan toward chain-of-frame reasoning, enabling dynamic frame generation during reasoning to provide visual evidence beyond the video frames for tackling more complex video reasoning tasks.
{
    \small
    \bibliographystyle{ieeenat_fullname}
    \bibliography{main}
}

\appendix
\clearpage
\setcounter{page}{1}
\maketitlesupplementary
\section{Related Work}
\subsection{Video Reasoning Tasks}
Recent advances in multimodal large language models (MLLMs) such as Qwen2.5-VL~\cite{Qwen2.5-VL}, Kimi-VL~\cite{kimiteam2025kimivltechnicalreport}, MiMo-VL~\cite{coreteam2025mimovltechnicalreport}, and GPT-4o~\cite{GPT-4o}, have substantially improved video understanding, including captioning~\cite{venugopalan2015translating}, question answering~\cite{vqaiccv}, and temporal grounding~\cite{gao2017talltemporalactivitylocalization}. However, these capabilities mainly reflect perceptual understanding~\cite{liu-etal-2024-tempcompass}, whereas video reasoning~\cite{li2024videovista}, which demands multi-hop deduction and causal inference across frames, remains insufficiently explored and evaluated. To address this gap, several benchmarks have been introduced to assess the reasoning capabilities of MLLMs, such as Video-Holmes~\cite{Cheng2025VideoHolmesCM}, VideoReasonBench~\cite{Liu2025VideoReasonBenchCM}, MMR-V~\cite{Zhu2025MMRVWL}, VRBench~\cite{Yu2025VRBenchAB}, and VCRBench~\cite{qi2025vcr}.
Unlike conventional video understanding tasks focused on recognizing visual content, these benchmarks require models to actively locate, connect, and reason over multiple relevant clues, thereby demanding a deeper comprehension of temporal dependencies and causal structures in dynamic visual narratives.

\subsection{Video Reasoning Models}
Inspired by the reasoning advancements of the DeepSeek-R1~\cite{deepseekai2025deepseekr1incentivizingreasoningcapability}, several studies~\cite{feng2025videor1reinforcingvideoreasoning,li2025videochatr1enhancingspatiotemporalperception} adopt reinforcement learning with verifiable rewards~\cite{deepseekai2025deepseekr1incentivizingreasoningcapability} (RLVR) to promote video reasoning in MLLMs. While these approaches~\cite{Ouyang2025SpaceRRM,feng2025videor1reinforcingvideoreasoning,li2025videochatr1enhancingspatiotemporalperception} encourage step-by-step reasoning, most are limited to text-only chain of thought, lacking explicit grounding in visual evidence, which often leads to unverified or hallucinated reasoning. To bridge this gap, concurrent works like Video-MTR~\cite{Xie2025VideoMTRRM} and FrameThinker~\cite{he2025framethinker} incorporate frame retrieval actions into the reasoning process, enabling dynamic evidence gathering and long-form understanding.
Despite great improvements in long video understanding~\cite{Zhou2024MLVUBM,wu2024longvideobench,wang2024lvbench}, these methods partially depend on benchmark-specific training sets and still lack a reliable evidence identification mechanism, rendering the retrieval actions less reliable. 
Motivated by this, we aim to develop a framework named Conan that incentivizes deductive-like reasoning abilities in MLLMs, combining precise evidence identification, logical multi-step reasoning, and confident action decision towards robust video reasoning.

\section{Dataset Construction}\label{App_dataset}
\noindent\textbf{Multi-scale Frame Categorization}. Frames are categorized into three types (\ie \textit{evidence}, \textit{context}, and \textit{distractor}), based on their frame-level scores in GenS-Video-150K~\cite{yao-etal-2025-generative}. Specifically, frames with scores below 3 are labeled as \textit{distractor}, those equal to 3 as \textit{context}, and those above 3 as \textit{evidence}.

\begin{figure*}[htb]
  \centering
  \begin{subfigure}{.9\linewidth}
    \includegraphics[width=\linewidth]{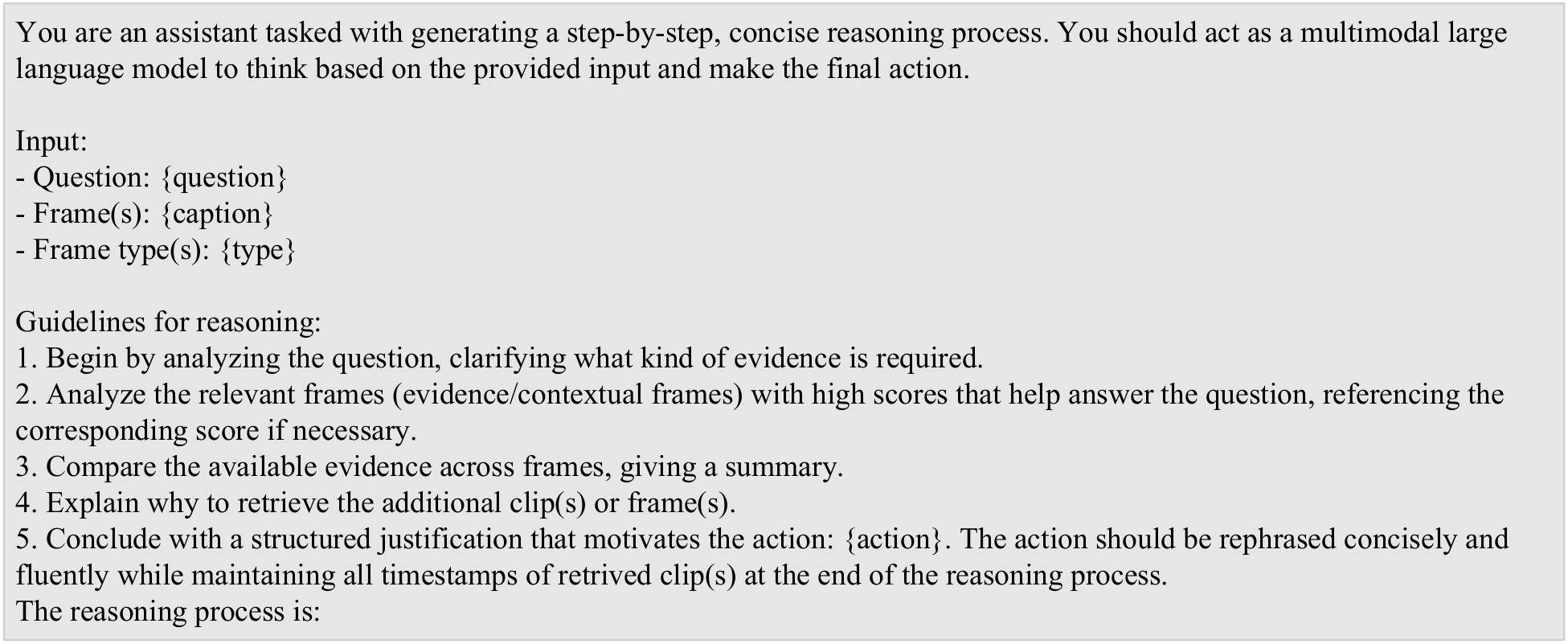}
    \caption{Prompt template for free-form question.}
  \end{subfigure}\hfill
  \begin{subfigure}{.9\linewidth}
    \includegraphics[width=\linewidth]{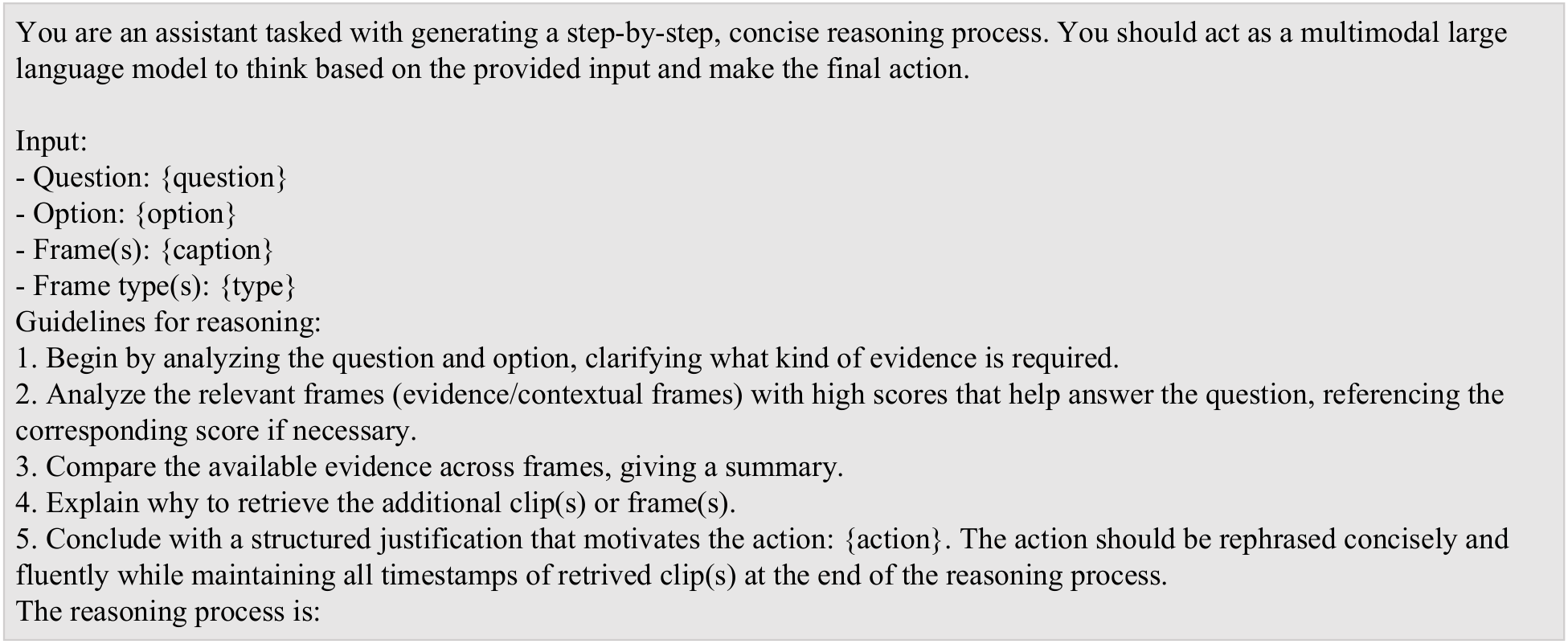}
    \caption{Prompt template for multi-choice question.}
  \end{subfigure}
  \caption{Prompt templates for reasoning trace construction.}
  \label{Fig:prompt}
\end{figure*}

\noindent\textbf{Reasoning Trace Construction}.
To establish the dynamic answering threshold, we adopt a double-verification mechanism. In particular, the accumulated frame descriptions and QA pair are first provided to K2 to assess whether the current evidence suffices to derive the correct answer. If so, K2 is prompted again without the answer to verify whether it can independently arrive at the correct option using the accumulated evidence. In addition, the prompt templates used to guide Kimi K2 in generating reasoning traces are illustrated in Figure~\ref{Fig:prompt}.

\begin{figure}[htb]
    \centering
    \includegraphics[width=0.98\linewidth]{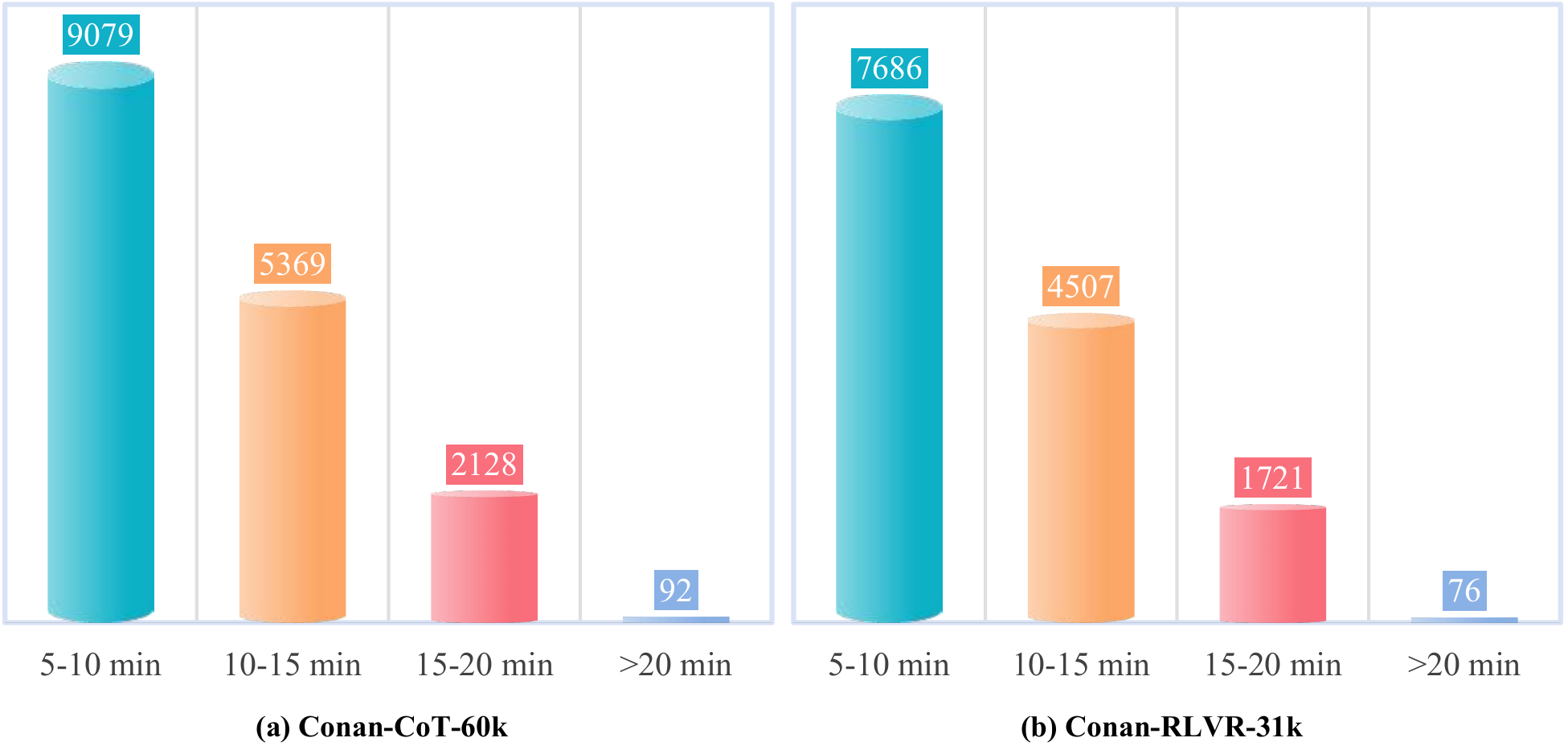}
     \caption{Duration distribution of (a) Conan-CoT-60k and (b) Conan-RLVR-31k.}
    \label{Fig:duration}
\end{figure}
\begin{figure}[htb]
    \centering
    \includegraphics[width=0.9\linewidth]{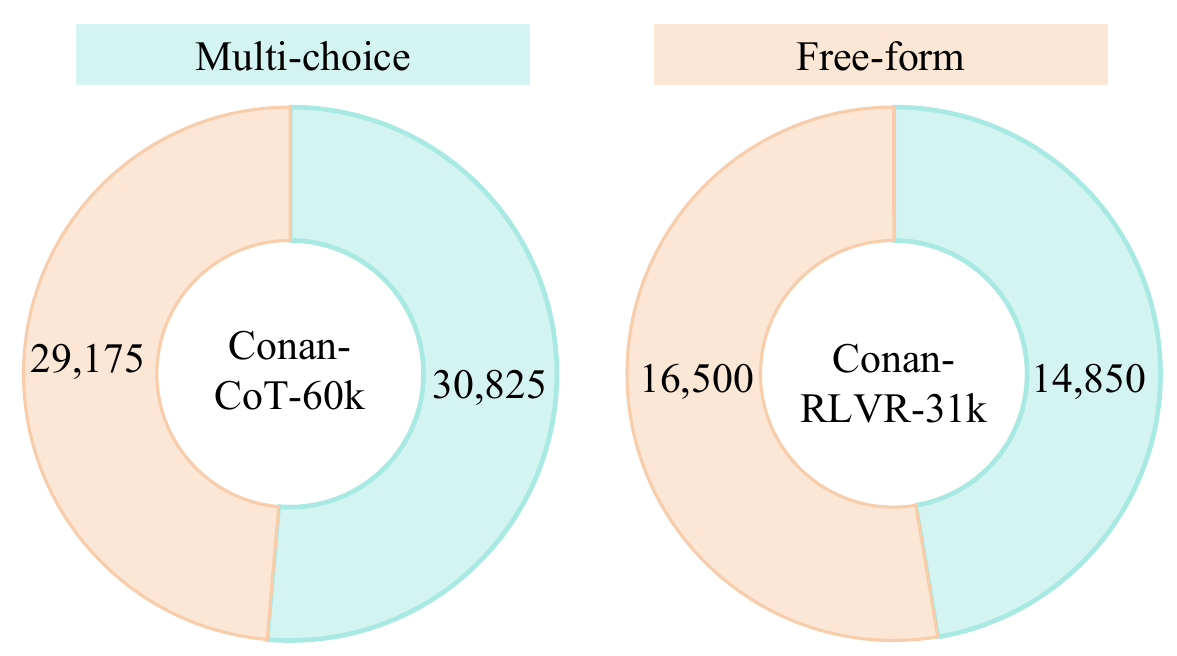}
     \caption{QA types distribution of Conan-CoT-60k and Conan-RLVR-31k.}
    \label{Fig:qa_distribution}
\end{figure}
\noindent\textbf{Dataset Statistics}. 
The Conan-91k dataset comprises $91,350$ samples, including $60,000$ for Conan-CoT-60k and $31,350$ for Conan-RLVR-31k. The duration distribution of each subset is shown in Figure~\ref{Fig:duration}, and the distribution of QA types is presented in Figure~\ref{Fig:qa_distribution}, respectively.

\section{Experiment}
\subsection{Benchmarks \& Baselines}\label{App_bench}
\noindent\textbf{Benchmark Descriptions.}
1) The six challenging video multi-step reasoning benchmarks.
\begin{itemize}
    \item \textbf{MMR-V}~\cite{Zhu2025MMRVWL}
    is a benchmark for multimodal deep reasoning in videos, containing $317$ videos and $1,257$ QA tasks designed to assess complex visual reasoning.
    \item \textbf{Video-Holmes}~\cite{Cheng2025VideoHolmesCM} evaluates high-level video reasoning across $1,837$ questions derived from $270$ manually annotated suspense short films, covering seven reasoning task types.
    \item \textbf{VRBench}~\cite{Yu2025VRBenchAB} is a long-form video reasoning benchmark with $960$ curated narrative videos spanning eight languages and seven categories, featuring $8,243$ human-labeled multi-step QA pairs targeting temporal and causal reasoning.
    \item \textbf{VCRBench}~\cite{qi2025vcr} assesses video chain-of-thought reasoning over $859$ videos with $1,034$ high-quality QA pairs. We adopt its multiple-choice subset for evaluation.
    \item \textbf{Human-P\&C}~\cite{li2025humanpcrprobingmllmcapabilities} includes over $6,000$ human-verified multiple-choice questions evaluating nine reasoning dimensions, covering both perceptual recognition and higher-level visual comprehension integrating commonsense or domain knowledge.
    \item \textbf{LongVideoReason}~\cite{chen2025longvila-r1} consists of $1,000$ long videos designed to comprehensively evaluate reasoning along four dimensions: temporal, goal-oriented, spatial, and narrative understanding.
    
\end{itemize}
The four long-video understanding benchmarks.
\begin{itemize}
     \item \textbf{LongVideoBench}~\cite{wu2024longvideobench} evaluates long-context multimodal video understanding with $6,678$ multiple-choice questions derived from videos up to one hour long, covering diverse real-world scenarios. We use its validation set and remove video subtitles for fair evaluation.
     \item \textbf{MLVU}~\cite{Zhou2024MLVUBM} is a multi-task benchmark containing $3,102$ questions across nine categories, specifically designed for long-video comprehension. We evaluate the multiple-choice subset from the dev split ($2,593$ samples).
     \item \textbf{LVBench}~\cite{wang2024lvbench} assesses models' ability to understand and extract information from long videos of up to two hours, comprising $1,549$ QA pairs in total.
     \item \textbf{Video-MME}~\cite{fu2024video} is a comprehensive benchmark for general video understanding, including $900$ videos and $2,700$ high-quality multiple-choice questions across diverse scenarios. Subtitles are excluded during evaluation for consistency.
\end{itemize}
\noindent\textbf{Baseline Descriptions.}
\begin{itemize}
    \item \textbf{GPT-4o}~\cite{GPT-4o} is a state-of-the-art MLLM developed by OpenAI, demonstrating strong performance across diverse vision-language tasks.
    \item \textbf{Gemini 1.5 Pro, Gemini 2.0 Flash} are advanced models from Google's Gemini family, achieving leading performance on video understanding. Particularly, Gemini 2.0 Flash exhibits improved complex reasoning capability.
    \item \textbf{LLaVA-OneVision-7B}~\cite{li2024llava} integrates the Qwen2~\cite{qwen2.5} language backbone with the SigLIP~\cite{DBLP:conf/iccv/ZhaiM0B23} vision encoder, achieving strong open-source performance in fine-grained visual understanding.
    \item \textbf{Kimi-VL-A3B-Instruct}~\cite{kimiteam2025kimivltechnicalreport}
    is an efficient Mixture-of-Experts (MoE) MLLM built upon the Moonlight~\cite{DBLP:journals/corr/abs-2502-16982} LLM and the high-resolution MoonViT~\cite{kimiteam2025kimivltechnicalreport} vision encoder.
    \item \textbf{InternVL3-8B}~\cite{chen2024expanding} employs the InternViT-300M-448px-V2\_5~\cite{chen2024internvl} vision encoder and Qwen2.5-7B~\cite{qwen2.5} backbone, delivering competitive open-source performance.
    \item \textbf{Video-R1}~\cite{feng2025videor1reinforcingvideoreasoning} enhances Qwen2.5-VL-7B-Instruct with CoT SFT and T-GRPO-based RLVR to improve video reasoning.
    \item \textbf{VideoChat-R1}~\cite{li2025videochatr1enhancingspatiotemporalperception} applies RLVR to Qwen2.5-VL-7B-Instruct, achieving strong spatiotemporal reasoning performance.
    \item \textbf{Video-MTR}~\cite{Xie2025VideoMTRRM} introduces multi-turn reasoning and a gated bi-level reward mechanism to enhance long-video understanding of Qwen2.5-VL-7B-Instruct.
     \item \textbf{Rewatch-R1}~\cite{zhang2025rewatch} augments RLVR with an Observation \& Reasoning reward to improve multi-step visual reasoning based on Qwen2.5-VL-7B-Instruct.
     \item \textbf{FrameThinker}~\cite{he2025framethinker} interleaves textual reasoning with visual frames, enabling multimodal chain-of-thought reasoning for long video understanding built upon Qwen2.5-VL-7B-Instruct. We do not include this model in the main results table, as it is not open-sourced prior to our submission.
    \item \textbf{Qwen2.5-VL-3B-Instruct, Qwen2.5-VL-7B-Instruct, Qwen2.5-VL-72B-Instruct}~\cite{Qwen2.5-VL} are part of the Qwen2.5-VL series, which combine the Qwen2.5~\cite{qwen2.5} language model with a redesigned Vision Transformer (ViT) architecture for enhanced visual grounding and understanding.
\end{itemize}

\subsection{More comparison results \& analyses}\label{App_exp}
\begin{table}[ht]
    \centering
    \resizebox{\linewidth}{!}{
        \begin{tabular}{l|c|c|c|c} 
        \toprule
             & \textbf{LongVideoBench} & \textbf{MLVU} & \textbf{LVBench} & \textbf{Video-MME}  \\ 
            \midrule
          \rowcolor{aliceblue}Qwen2.5-VL-7B-Instruct~\cite{Qwen2.5-VL}  
          &48.9&52.8&34.4&55.8 \\
          \midrule
          Video-R1~\cite{feng2025videor1reinforcingvideoreasoning} &55.6&62.5&38.3&58.6\\
          VideoChat-R1~\cite{li2025videochatr1enhancingspatiotemporalperception}&54.3&60.5&38.0&56.9\\
          \midrule
          Rewatch-R1~\cite{zhang2025rewatch} &50.5&55.2&37.2&58.9 \\
          Video-MTR~\cite{Xie2025VideoMTRRM} &56.4&59.7&38.6&59.3\\
          FrameThinker~\cite{he2025framethinker}&52.9&59.1&36.6&-\\
          \midrule
          \rowcolor{lightblue} Conan &\textbf{56.6}~\textcolor{jweigreen}{\small($\uparrow 7.7$ )}&\textbf{63.4}~\textcolor{jweigreen}{\small($\uparrow 10.6$ )}&\textbf{39.2}~\textcolor{jweigreen}{\small($\uparrow 4.8$ )}&\textbf{60.5}~\textcolor{jweigreen}{\small($\uparrow 4.7$ )}\\
        \bottomrule
        \end{tabular}
    }
    \caption{Evaluation results on long video understanding. The results of FrameThinker are taken from its original paper, which does not evaluate the entire Video-MME benchmark.}
    \label{Tab:longvideo_exp}
\end{table}
\noindent\textbf{Long Video Understanding.}
Beyond multi-step reasoning, Conan exhibits strong generalization to long video understanding tasks. As presented in Table~\ref{Tab:longvideo_exp}, Conan consistently outperforms Qwen2.5-VL-7B-Instruct across LongVideoBench~\cite{wu2024longvideobench}, MLVU~\cite{Zhou2024MLVUBM}, LVBench~\cite{wang2024lvbench}, and Video-MME~\cite{fu2024video}, achieving state-of-the-art performance compared with both text-CoT and video-CoT models. These results indicate that the high-quality, multi-scale evidence reasoning data and progressive training curriculum not only enhance multi-step reasoning but also effectively boost long video understanding.

\begin{figure}[htb]
    \centering
    \includegraphics[width=0.98\linewidth]{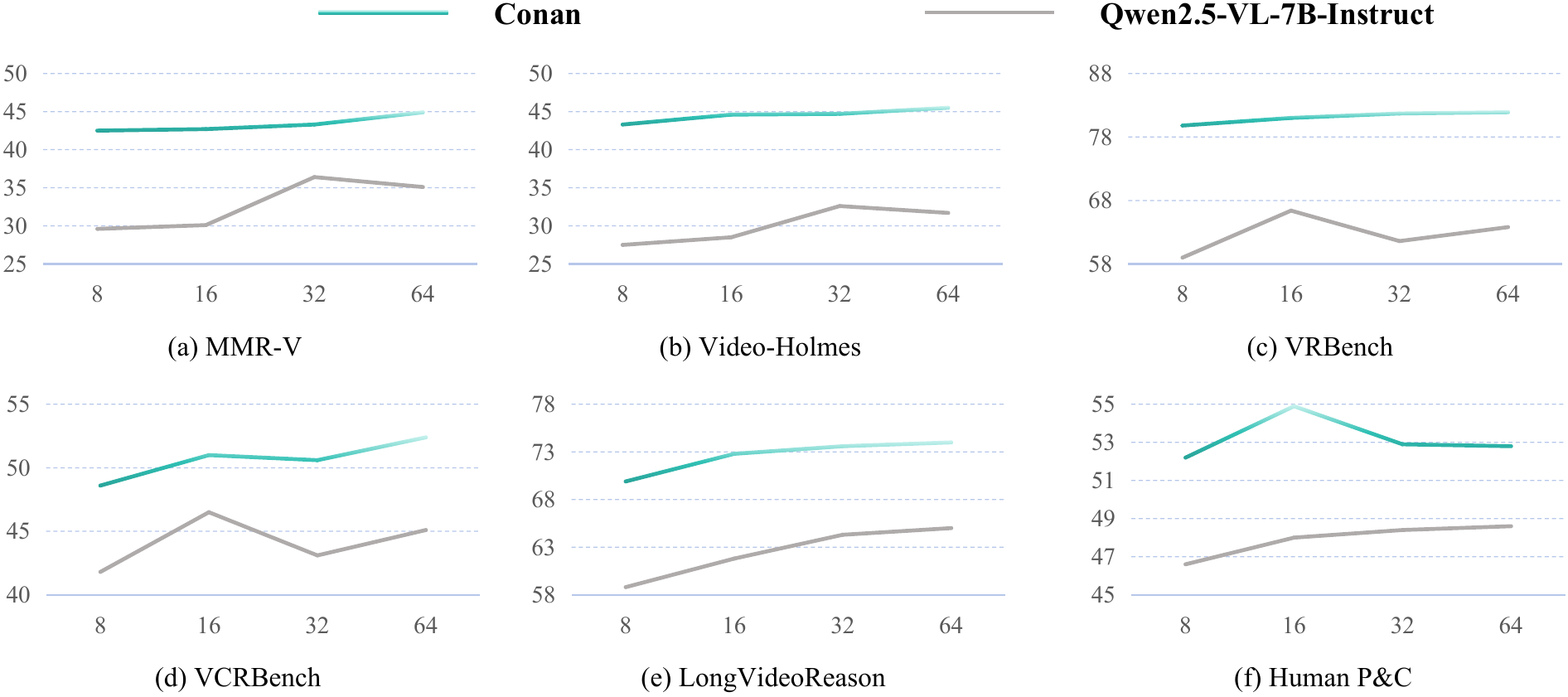}
     \caption{Performance variations of Conan and Qwen2.5-VL-7B-Instruct across different input frame numbers.}
    \label{Fig:efficiency}
\end{figure}
\begin{table*}[ht]
    \centering
    \resizebox{\textwidth}{!}{
        \begin{tabular}{l|c|c|c|c|c|c|c|c}
        \toprule
            & \#Params  & \textbf{MMR-V} & \textbf{Video-Holmes} & \textbf{VRBench} & \textbf{VCRBench$\ast$}&  \textbf{Human-P\&C} &\textbf{LVR} & \textbf{Overall}  \\
        \midrule
        Qwen2.5-VL-3B-Instruct~\cite{Qwen2.5-VL}
           &3B&29.0 &28.8 &59.7 &38.0 &41.7 &57.6&42.5   \\
        \rowcolor{lightblue} Conan-3B& 3B 
        & {33.1}~\textcolor{jweigreen}{ \small($\uparrow 4.1$ )}    &{31.9}~\textcolor{jweigreen}{ \small($\uparrow 3.1$ )}  & {68.0}~\textcolor{jweigreen}{ \small($\uparrow 8.3$ )}  & {42.7}~\textcolor{jweigreen}{ \small($\uparrow 4.7$ )} & {44.3}~\textcolor{jweigreen}{ \small($\uparrow 2.6$ )} &{61.3}~\textcolor{jweigreen}{ \small($\uparrow 3.7$ )}&{46.9}~\textcolor{jweigreen}{ \small($\uparrow 4.4$ )} \\ 
        \midrule
           InternVL3-8B~\cite{chen2024expanding}
        &8B&35.5&32.8&75.8&45.7&51.0&68.0&51.5\\
        \rowcolor{lightblue} Conan-Intern-8B&8B&43.3~\textcolor{jweigreen}{ \small($\uparrow 7.8$ )}&41.8~\textcolor{jweigreen}{ \small($\uparrow 9.0$ )}&81.7~\textcolor{jweigreen}{ \small($\uparrow 5.9$ )}&48.4~\textcolor{jweigreen}{ \small($\uparrow 2.7$ )}&50.2~\textcolor{jweired}{ \small($\downarrow 0.8$ )}
        &70.8~\textcolor{jweigreen}{ \small($\uparrow 2.8$ )}&56.0~\textcolor{jweigreen}{ \small($\uparrow 4.5$ )}\\
        \bottomrule
        \end{tabular}
    }
    \caption{Performance comparison on multi-step reasoning of Conan variants and their corresponding base models. \textbf{Overall} represents the average results across six multi-step reasoning benchmarks. $\ast$ indicates the multiple-choice subset.}
    \label{Tab:app_main_exp}
\end{table*}
\noindent\textbf{Frame Efficiency Analysis.}
We compare the efficiency of Conan with Qwen2.5-VL-7B-Instruct under varying numbers of initial input frames. As shown in Figure~\ref{Fig:efficiency}, Conan with only $8$ initial input frames outperforms the base model that uses $64$ frames, demonstrating its superior efficiency and accuracy in visual reasoning. However, we observe a slight performance decline on the Human-P\&C benchmark as the number of input frames increases. This is reasonable, as introducing more initial frames also introduces additional distractors that are irrelevant to answering the question, thereby increasing the difficulty of frame identification.
\begin{figure*}[htb]
    \centering
    \includegraphics[width=0.9\textwidth]{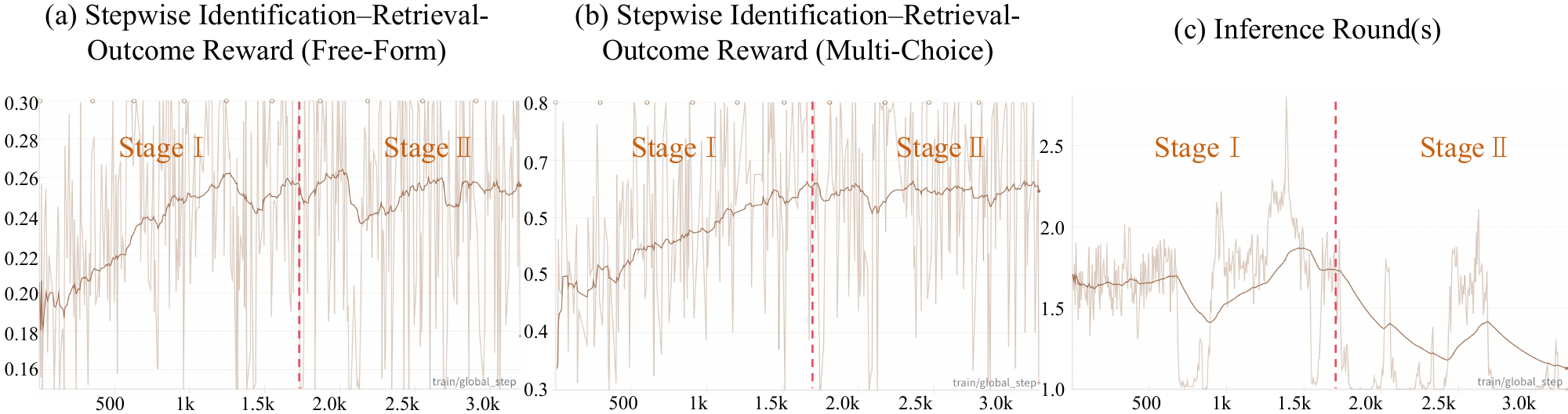}
     \caption{Training dynamics in the AIR RLVR process of Conan.}
    \label{Fig:training_dynamics}
\end{figure*}
\\ \noindent\textbf{Generalizability across model sizes and architectures.}
To evaluate the generalizability of our dataset and training framework, we apply them to different base models: Qwen2.5-VL-3B-Instruct and InternVL3-8B. As shown in Table~\ref{Tab:app_main_exp}, both Conan-3B and Conan-Intern-8B achieve consistent performance improvements over their respective baselines, demonstrating the strong extensibility and scalability of the proposed Conan-91k dataset and two-phase training framework across varying model sizes and architectures. 
Notably, although InternVL3-8B demonstrates stronger baseline performance than Qwen2.5-VL-7B-Instruct, Conan-Intern-8B underperforms Conan-7B in multi-step reasoning. This discrepancy likely arises because InternVL3-8B employs a perception-oriented visual encoder and a relatively shallow cross-modal fusion module, which constrain its ability to integrate multi-step temporal cues and causal dependencies during reasoning~\cite{chen2024expanding}. In contrast, Qwen2.5-VL-7B-Instruct inherits reasoning priors and instruction-following capabilities from its text-based foundation model, thereby facilitating more effective compositional and multi-hop reasoning~\cite{Qwen2.5-VL}.

\subsection{Training Dynamics of AIR RLVR}
To gain a deeper understanding of how the model's behavior evolves during end-to-end reinforcement learning, we perform a fine-grained analysis of its training dynamics. As shown in Figure~\ref{Fig:training_dynamics}, the training process can be divided into two stages:

\noindent\textbf{Stage I: Accuracy-Oriented Evidence Exploration.} 
Building upon the solid foundation established through multi-stage progressive cold-start, the model initially enters a phase characterized by frequent yet progressively more accurate frame retrieval. During this stage, it actively queries additional clips to maximize the joint identification–retrieval–outcome reward across tasks. The steadily rising reward curve indicates that the model adopts an exploratory strategy: Frequently engaging with the visual context to accumulate sufficient evidence for accurate reasoning. This stage marks a transitional period in which the model learns the value of comprehensive evidence collection before optimizing retrieval efficiency.

\noindent\textbf{Stage II: Efficient Evidence Retrieval.} 
As training progresses, the model transitions to a more refined and selective retrieval policy. It significantly reduces retrieval frequency while maintaining high reward accuracy, indicating that Conan has internalized a compact and efficient multi-step reasoning strategy, retrieving evidence only when necessary, much like a detective who strategically gathers key clues rather than exhaustively examining all information.
\begin{figure*}[htb]
  \centering
  \begin{subfigure}{.9\linewidth}
    \includegraphics[width=\linewidth]{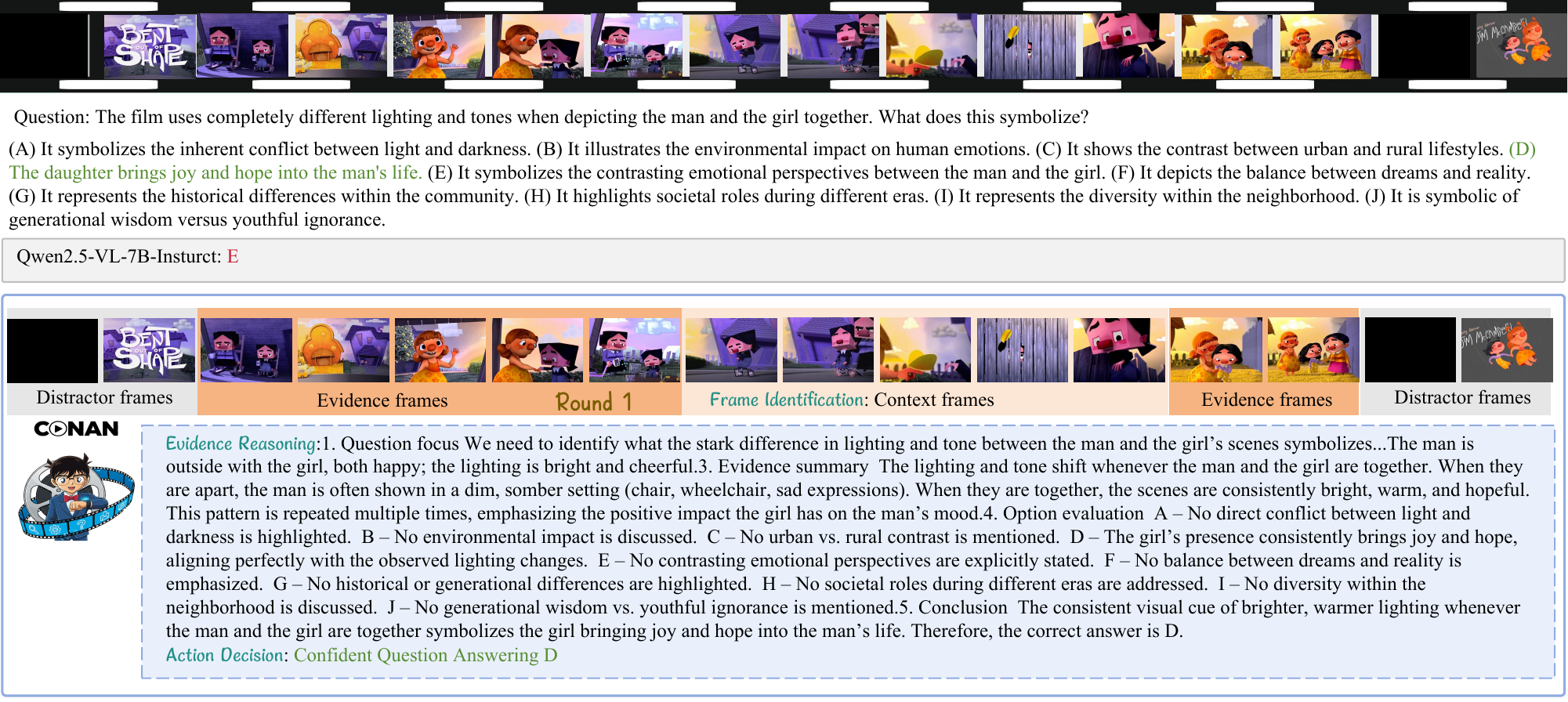}
    \caption{A qualitative example from MMR-V.}
  \end{subfigure}\hfill
  \begin{subfigure}{.9\linewidth}
    \includegraphics[width=\linewidth]{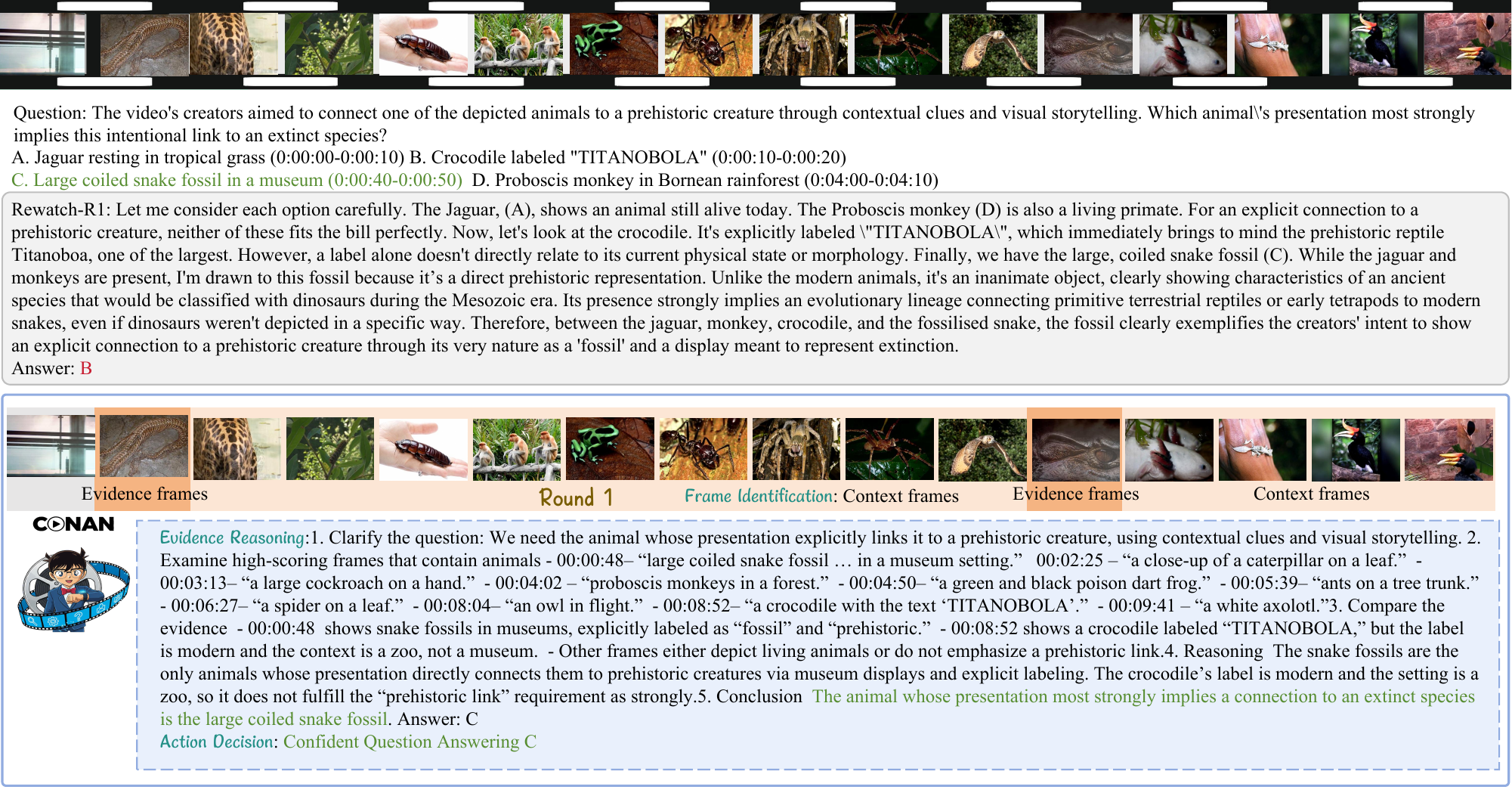}
    \caption{A qualitative example from LongVideoReason.}
  \end{subfigure}
  \caption{Two qualitative examples from MMR-V and LongVideoReason are presented, illustrating the reasoning traces and final answers of Rewatch-R1 and Conan for comparison.}
  \label{Fig:case2}
\end{figure*}
\subsection{More qualitative analyses}\label{App_qualitative}
As shown in Figure~\ref{Fig:case2}, in case (a), while Qwen2.5-VL-7B-Instruct incorrectly selects option ``E'' by relying on superficial textual associations, Conan correctly identifies option ``D'' through structured, evidence-grounded reasoning. During the \textit{Frame Identification} stage, Conan identifies context and evidence frames that emphasize the contrast in lighting and tone between the man and the girl. It then performs explicit evidence reasoning, linking the recurring visual pattern, bright, warm lighting accompanying the girl's appearance, to the symbolic themes of joy and hope she brings to the man's life. Finally, Conan synthesizes this reasoning into a confident and accurate conclusion.
In case (b), despite Rewatch-R1 producing a plausible reasoning process, it fails to anchor its analysis in explicit visual evidence and ultimately predicts the wrong answer due to misalignment between reasoning (answer C) and action (answer B). In contrast, Conan accurately identifies the correct option through structured reasoning grounded in evidence frames. In the \textit{Frame Identification} stage, Conan localizes key frames featuring the fossil exhibit, then performs evidence reasoning to associate the fossil's visual features and its museum context, explicitly symbolizing extinction, with the concept of a prehistoric creature. Conan then integrates this reasoning into a coherent and confident response, showcasing robust visual grounding and strong reasoning–action alignment.
Both of the two cases highlight Conan's ability to perform compositional reasoning over visual cues, aligning symbolic interpretation with concrete visual evidence rather than relying on spurious correlations.
\begin{figure*}[htb]
  \centering
  \begin{subfigure}{.9\linewidth}
    \includegraphics[width=\linewidth]{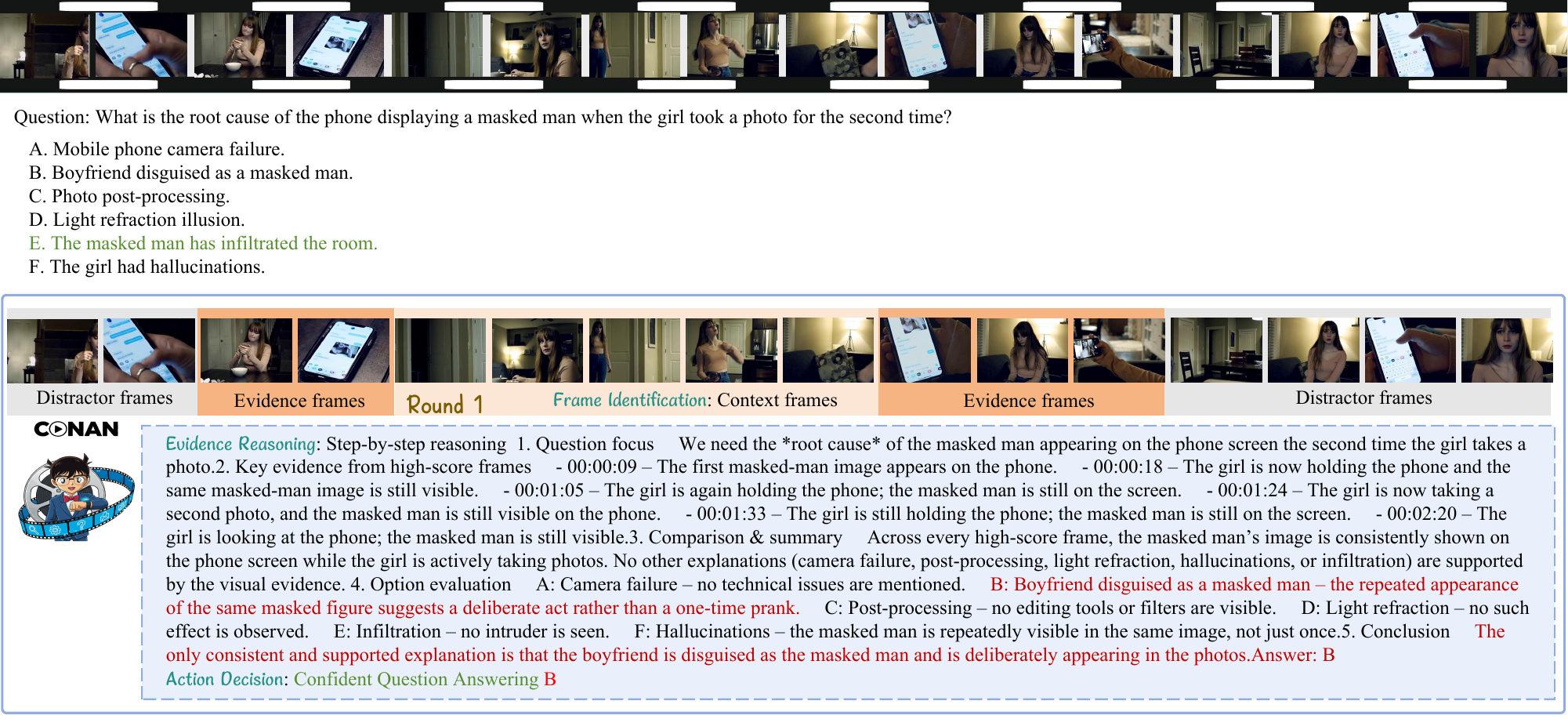}
    \caption{Error case from Video-Holmes.}
  \end{subfigure}\hfill
  \begin{subfigure}{.9\linewidth}
    \includegraphics[width=\linewidth]{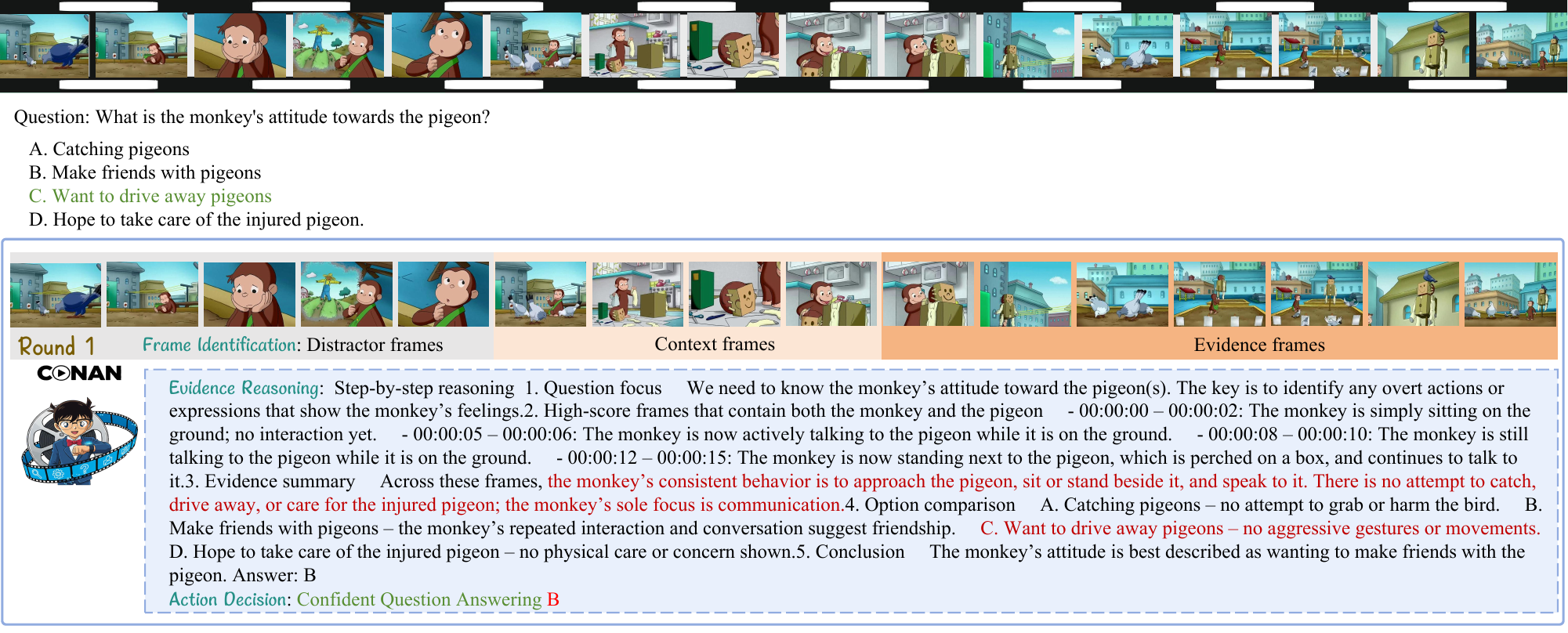}
    \caption{Error case from VCRBench.}
  \end{subfigure}
  \caption{Two error cases of Conan from Video-Holmes and VCRBench.}
  \label{Fig:error_case}
\end{figure*}
\\
\noindent\textbf{Error Analyses.}
Figure~\ref{Fig:error_case} presents two representative error cases from Video-Holmes and VCRBench. In error case (a), Conan accurately identifies and aligns the key evidence frames showing the masked man's repeated appearance on the phone screen. However, it incorrectly infers that the masked man is the boyfriend in disguise (\textit{Option B}) rather than recognizing the correct explanation of supernatural infiltration (\textit{Option E}). This misjudgment arises from a hallucinated causal link, interpreting visual recurrence as intentional disguise, while neglecting contextual cues, such as the absence of any indication of the boyfriend's presence. The case reveals Conan's limitation in distinguishing plausible narrative causality from repetitive visual patterns.
In error case (b), Conan interprets the monkey's seemingly gentle behavior toward the pigeon as an intention to befriend it (\textit{Option B}). While the model accurately focuses on consistent proximity and interaction frames, it overlooks subtle affective cues, such as the monkey's frustrated expressions and its act of using a scarecrow to drive the pigeon away, which point to the correct answer (\textit{Option C}). This error underscores Conan's limitation in capturing fine-grained emotional and social dynamics, leading to an over-reliance on surface-level behavioral patterns. 

Collectively, these cases reveal Conan's ongoing challenges in integrating implicit visual evidence and affective reasoning beyond superficial visual and temporal correlations.

\end{document}